\documentclass{article}

\PassOptionsToPackage{numbers, compress}{natbib}

\usepackage[preprint]{neurips_2025}
 \usepackage{amsmath}
\usepackage{fontawesome5}
\usepackage{wrapfig}
\usepackage[table,xcdraw]{xcolor}
\usepackage{multirow}
\usepackage[utf8]{inputenc} 
\usepackage[T1]{fontenc}    
\usepackage{hyperref}       
\usepackage{url}            
\usepackage{booktabs}       
\usepackage{amsfonts}       
\usepackage{nicefrac}       
\usepackage{microtype}      
\usepackage{xcolor}         
\usepackage[most]{tcolorbox}
\usepackage{fontawesome5}
\usepackage{hyperref}
\usepackage{array} 
\usepackage{listings}
\usepackage{graphicx}  
\usepackage{subcaption} 
\usepackage{float}      
\usepackage{fontawesome5}
\definecolor{lightblue}{RGB}{235,245,255} 
\definecolor{liautoblue}{RGB}{71,111,182} 
\definecolor{textred}{RGB}{128,0,0}
\usepackage{titlesec}
\titleformat{\section}
  {\large\bfseries\color{liautoblue}}{\thesection}{1em}{}
\titleformat{\subsection}
  {\bfseries\color{liautoblue}}{\thesubsection}{1em}{}
  \titleformat{\subsubsection}
  {\bfseries\color{liautoblue}}{\thesubsubsection}{1em}{}
\usepackage[most]{tcolorbox}
\newtcolorbox{liautoabstract}{
    colback=lightblue,
    colframe=white,
    boxrule=0pt,
    arc=2mm,
    left=4mm,
    right=4mm,
    top=5mm,
    bottom=5mm,
    enhanced, 
    before upper={\setlength{\parindent}{0pt}} 
}
\usepackage[most]{tcolorbox} 
\usepackage{xeCJK}

\newtcolorbox{stepTitle}[1]{
    enhanced,
    colback=gray!5,    
    colframe=black!50,  
    boxrule=-1pt,
    arc=0mm,            
    left=2mm, right=2mm, top=1mm, bottom=1mm,
    fontupper=\small,  
    title=#1
}

\tcbset{
    stepbox/.style={
        enhanced,
        colback=gray!20,         
        colframe=gray!50,        
        boxrule=0.5pt,
        title={#1},              
        fonttitle=\bfseries,     
        left=2mm, right=2mm, top=1mm, bottom=1mm
    }
}

\newtcolorbox[auto counter]{case}[2][]{ 
    enhanced,
    colback=gray!5,
    colframe=black!70,
    coltitle=white,
    fonttitle=\bfseries\sffamily,
    fontupper=\small,
    arc=1.5mm,
    boxrule=0.5pt,
    title=Case study \thetcbcounter: #2, 
    left=1mm, right=1mm, top=2mm, bottom=2mm,
    label type=case, 
    #1               
}

\newtcolorbox{toolbox}[1]{
    enhanced,                 
    colback=gray!5,           
    colframe=black!70,        
    coltitle=white,           
    fonttitle=\bfseries\sffamily,
    fontupper=\small,
    arc=1.5mm,                
    boxrule=0.5pt,            
    title=#1,                 
    left=1mm, right=1mm, top=2mm, bottom=2mm
}

\usepackage{fancyhdr}
\usepackage{graphicx} 
\hypersetup{
    colorlinks=true,
    urlcolor=liautoblue,
}

\title{MindWatcher: Toward Smarter Multimodal Tool-Integrated Reasoning}

\author{%
  MindGPT-ov Team \\
   Li Auto Inc   \\
}

\begin{document}

\maketitle

\begin{liautoabstract} 

Traditional workflow-based agents exhibit limited intelligence when addressing real-world problems requiring tool invocation. Tool-integrated reasoning\,(TIR) agents capable of autonomous reasoning and tool invocation are rapidly emerging as a powerful approach for complex decision-making tasks involving multi-step interactions with external environments.
In this work, we introduce MindWatcher, a TIR agent integrating interleaved thinking and multimodal chain-of-thought (CoT) reasoning. MindWatcher can autonomously decide whether and how to invoke diverse tools and coordinate their use, without relying on human prompts or workflows. The interleaved thinking paradigm enables the model to switch between thinking and tool calling at any intermediate stage, while its multimodal CoT capability allows manipulation of images during reasoning to yield more precise search results.
We implement automated data auditing and evaluation pipelines, complemented by manually curated high-quality datasets for training, and we construct a benchmark, called MindWatcher-Evaluate Bench\,(MWE-Bench), to evaluate its performance. MindWatcher is equipped with a comprehensive suite of auxiliary reasoning tools, enabling it to address broad-domain multimodal problems. A large-scale, high-quality local image retrieval database, covering eight categories including cars, animals, and plants, endows model with robust object recognition despite its small size. Finally, we design a more efficient training infrastructure for MindWatcher, enhancing training speed and hardware utilization.
Experiments not only demonstrate that MindWatcher matches or exceeds the performance of larger or more recent models through superior tool invocation, but also uncover critical insights for agent training, such as the genetic inheritance phenomenon in agentic RL.
The agent reasoning framework, MWE-Bench, three smaller-scale agent models\,(2B, 3B, and 4B) distilled from MindWatcher 32B, and related resources will be open-sourced.

\vspace{3mm}
    {\color{liautoblue!30}\rule{\linewidth}{0.5pt}} 
    \vspace{2mm}

    \small 
    \renewcommand{\arraystretch}{1.3} 
\begin{tabular}{@{} l l @{}}
        {\faCalendar*} & \textbf{Last Update Date:} December 29, 2025 \\
        {\color{liautoblue}\faEnvelope} & \textbf{Correspondence:} {chenwei10@lixiang.com} \\
        {\faGithub} & \textbf{Code:} \href{https://github.com/TIMMY-CHAN/MindWatcher}{https://github.com/TIMMY-CHAN/MindWatcher} \\
        {\raisebox{-0.5ex}{\includegraphics[height=1.1em]{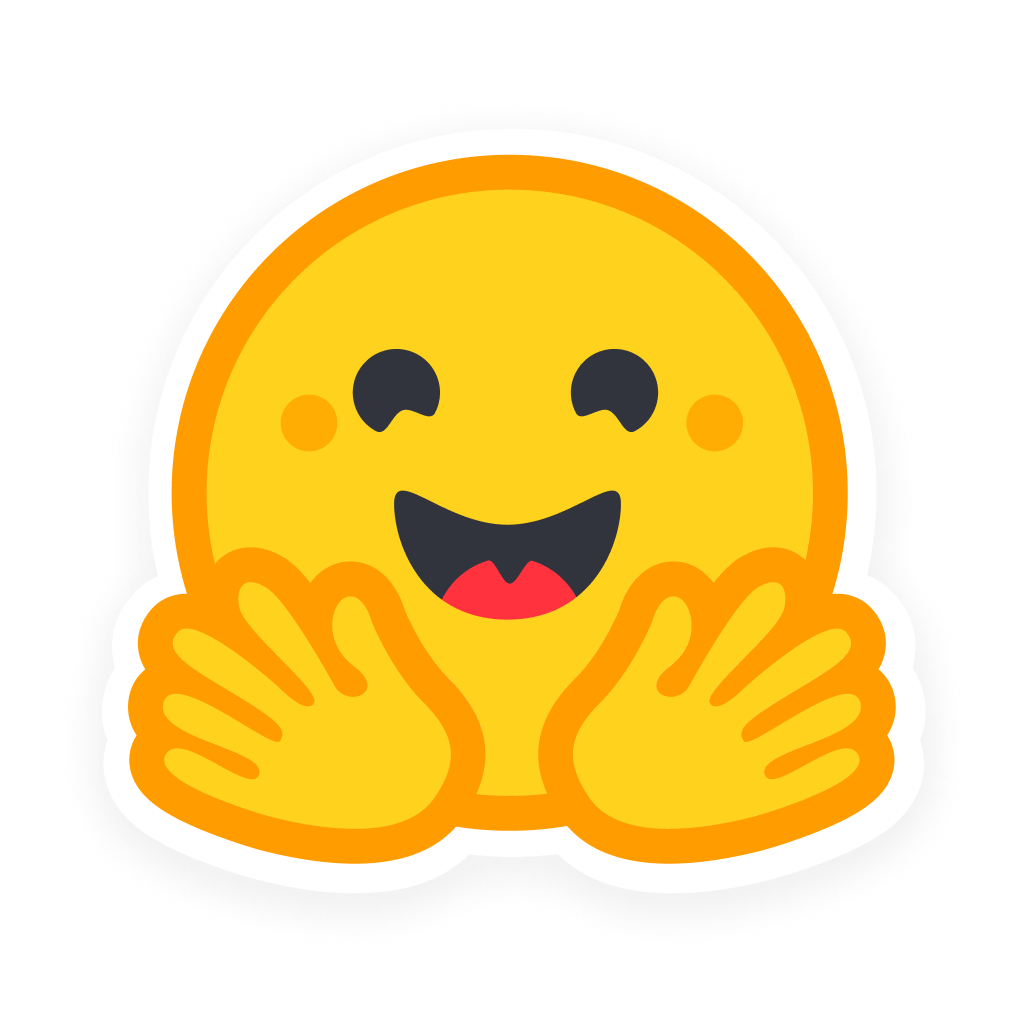}}} & \textbf{Hugging Face:} \href{https://huggingface.co/datasets/Lost-Cloud/MWE-Bench}{https://huggingface.co/datasets/Lost-Cloud/MWE-Bench}
    \end{tabular}\end{liautoabstract}

\section{Introduction}
\begin{figure}
    \centering
    \includegraphics[trim={0cm 1cm 0cm 1cm}, clip, width=\textwidth]{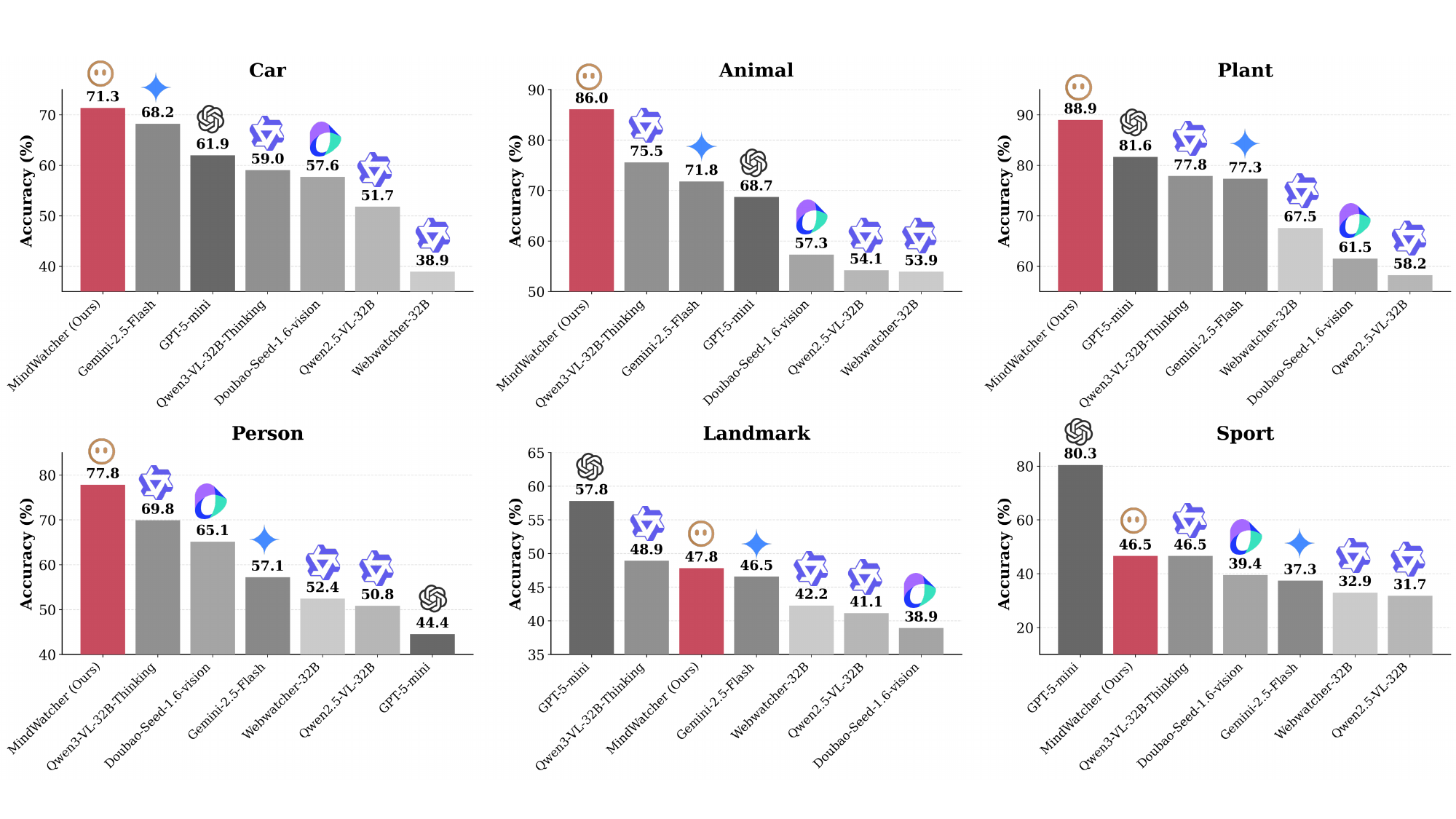}
    \caption{MWE-Bench Performance of MindWatcher.}
    \label{fig1}
\end{figure}

Large language models (LLMs)~\cite{gpt, GPT-4-REPORT, deepseek-r1, yang2024qwen2.5, gemini2.5, llama,mindgpt-4ov} have achieved remarkable progress in recent years, demonstrating strong capabilities in language understanding, knowledge acquisition, and complex reasoning tasks. However, despite the powerful world knowledge and multimodal capabilities of the latest models such as Gemini 2.5 Pro~\cite{gemini2.5}, most LLMs remain fundamentally constrained by the limits of their parametric knowledge: they struggle to cover long‑tail information and fine‑grained domain‑specific knowledge~\cite{chen2024detecting}, and they cannot directly access real‑time information that emerges after training. These structural bottlenecks hinder their reliability in many real‑world applications, especially those requiring external knowledge, multi‑step information integration, or cross‑modal reasoning. Equipping LLMs with external tools has therefore become a key strategy to overcome these limitations. By connecting models with retrieval engines~\cite{RAG,  Shao_2023_CVPR}, computation tools, or code interpreters, the boundary of problem‑solving capabilities can be substantially extended.

Traditional tool‑augmented approaches typically rely on manually designed workflows~\cite{khattab2022demonstrate, shi2025flowagent} to orchestrate tool invocation, yet such methods exhibit limited adaptability when confronted with the diversity and uncertainty inherent in open-domain environments, which become even more fragile when handling cross‑modal demands. Multi‑agent systems~\cite{autogen, shen2023hugginggpt, li2025webweaver, li2025mccd} partially alleviate these issues: a powerful planner agent is responsible for decision‑making, while tool‑specialized agents execute designated subtasks. This architecture has become highly popular in the industry and significantly improves system flexibility and scalability. However, it also introduces new complexity and overhead, including redundant model deployment and latency caused by chained interactions, which limits its further expansion. With the emergence of thought‑augmented models~\cite{cot, deepseek-r1}, the research community increasingly recognizes that intelligent systems need not rely on multi‑component designs: a single unified language model can assume both planning and acting roles. This has led to the rise of Tool‑Integrated Reasoning (TIR) methods~\cite{qiao2025webresearcher}, exemplified by the ReAct~\cite{yao2022react} paradigm. The core idea is to let the model explicitly generate intermediate thoughts, autonomously invoke tools, and iteratively make decisions based on environmental feedback. TIR agents can dynamically plan multi‑step operations in open‑world tasks and achieve end‑to‑end problem solving, making them a promising path toward more general‑purpose agents. 

However, current TIR systems still fall short of being truly practical and general intelligent agents, with significant limitations across several key dimensions. From an application perspective, existing TIR agents~\cite{qiao2025webresearcher, li2025webweaver, tool-star} are predominantly focused on text‑based tasks, particularly DeepSearch‑style reasoning centered on retrieval. Only a small number of works~\cite{geng2025webwatcher, zheng2025deepeyes} attempt to introduce visual capabilities, and most rely solely on image search tools without enabling the agent to directly manipulate images or perform fine‑grained cross‑modal reasoning to support problem solving. This severely limits their performance on multimodal tasks and prevents them from tackling the many vision‑driven decision‑making scenarios found in real‑world environments. 

From a training methodology perspective, TIR agents face a triple challenge across data, algorithms, and training frameworks. High‑quality reasoning trajectories involving multiple tools and multi‑step interactions are extremely difficult to construct manually. SFT‑based training~\cite{instructgpt,chen2024efficiency} often causes models to “imitate” the thought‑action format rather than truly “learn” the underlying strategy—manifested in excessive, redundant tool calls on simple problems and substantial performance degradation on general benchmarks. Moreover, existing training frameworks lack fine‑grained supervision over the interleaved process of thinking, tool invocation, and subsequent reasoning, preventing models from forming stable and reliable tool‑use behaviors and exacerbating issues such as tool misuse and unnecessary calls. From the perspective of tool ecosystems, many core retrieval capabilities, especially visual retrieval, rely on expensive external APIs. Their high cost under frequent invocation further constrains the practical deployment of TIR agents in local or enterprise settings.

To address the challenges outlined above, we introduce MindWatcher, a TIR agent capable of autonomous planning and execution, multimodal perception, and robust tool coordination. Leveraging an interleaved thinking paradigm and a multimodal Chain‑of‑Thought\,(CoT) mechanism, MindWatcher can flexibly alternate between internal thinking and external tool invocation at any stage of the reasoning process. By integrating fine‑grained visual operations into the reasoning chain, the agent achieves precise region‑level visual perception and more accurate cross‑modal information retrieval.

To avoid the drawbacks of conventional SFT, such as rigid imitation of reasoning formats and redundant tool calls on simple tasks, MindWatcher abandons standard SFT and instead adopts a continuous reinforcement learning\,(RL) strategy conducted in both real and offline environments. We develop two automated image–text pair construction pipelines to reduce data generation costs. In parallel, we equip MindWatcher with a comprehensive set of tools that cover core multimodal reasoning needs, including image region cropping and zooming, object grounding and visual search, external text retrieval, webpage content extraction, and local Python code interpreter. Moreover, we construct a large‑scale local visual corpus spanning categories such as person, animals, plants, cars, landmarks, and logos. We also build a new multimodal benchmark: MindWatcher-Evaluate Benchmark (MWE-Bench) for systematically evaluating agentic multimodal tool‑use and reasoning capabilities.

At the system level, we design an RL training pipeline that supports asynchronous tool invocation, significantly improving learning efficiency. We also introduce a new GRPO-based agentic RL algorithm, which introduce step-wise normalization which ensure the optimization objective on individual action segments rather than the global token stream.

As shown in Figure~\ref{fig1}, MindWatcher demonstrates strong generality and efficiency across a wide range of tasks on MWE-Bench. The 32B model achieves state‑of‑the‑art\,(SOTA) performance in tool‑augmented reasoning while maintaining robust general capabilities, and we distilled and open-sourced 2B, 3B, and 4B variants based on the MindWatcher, which also exhibit highly competitive results.

\section{Method}
\begin{figure}
    \centering
    \includegraphics[width=1\linewidth]{image/kuajiatuv2.pdf}
    \caption{\textbf{The Working Paradigm of MindWatcher.} To address complex multimodal question answering tasks, we train our model using continuous RL to develop Multimodal CoT capabilities. By integrating interleaved thinking, the model is able to interact with the environment and autonomously invoke tools in the toolbox. Furthermore, to facilitate more accurate and lowcost visual search, MindWatcher have constructed a large-scale local retrieval corpus spanning eight major categories. }
    \label{fig2}
\end{figure}
\subsection{Working Paradigm}

To support flexible multimodal reasoning and autonomous tool use, MindWatcher models the TIR process as a Markov Decision Process (MDP). Given an initial user prompt $s_0$, the agent interacts with the environment by generating an interleaved sequence of actions and tool‑grounded observations:
\begin{equation}
    Y = \{a_0, obs_0, a_1, obs_1, \dots, obs_{n-1}, a_n\}.
\end{equation}
Each action $a_j$ is executed against the environment—typically through a thinking process and a tool call—yielding an observation $obs_j$, which is appended to the context and becomes part of the next state. The agent iteratively continues this process until generating the final action $a_n$, which contains the concluding response. 

\textbf{Interleaved Thinking and Multimodal CoT}
MindWatcher implements this MDP through an autoregressive generation loop. At each step $t$, the Policy $\pi_\theta(a_t | s_t)$ (parameterized by the LLM) conditions on the full history $s_t$.
Distinct from traditional approaches where actions are strictly physical tool calls, we define a unified action space $\mathcal{A} = \mathcal{A}_{thought} \cup \mathcal{A}_{tool}$.
In implementation, thoughts and tool calls are serialized through dedicated <think>\dots<\textbackslash think> and <tool\_call>\dots<\textbackslash tool\_call> tags, enabling the model to interleave reasoning and action generation within a single decoding sequence. MindWatcher further incorporates a multimodal CoT~\cite{cot, ComT} mechanism, which allows the agent to “think with images” by embedding image‑dependent operations into its reasoning chain.

As shown in Figure~\ref{fig2}, given an image and a complex query, the model enters an iterative planning and tool-call process. After each tool-call completes, the tool response of current stage is obtained. Subsequently, the next action is determined based on the tool response, ultimately yielding the query result.
\subsection{Training Algorithm}

While SFT remains the prevailing paradigm for training TIR agents, our empirical observations reveal significant limitations. We found that fine-tuning already robust instruction-following or thinking models on trajectory data often incurs a heavy "alignment tax", severely degrading performance on general-purpose tasks. Furthermore, SFT tends to induce \textit{tool abuse}, manifested as redundant invocations for trivial queries and excessive, ineffective looping in complex scenarios. Consequently, we adopt a pure RL approach to endow MindWatcher with genuine decision-making and self-correction capabilities.

\subsubsection{Step-wise Normalized GRPO}

We employ an enhanced version of Group Relative Policy Optimization (GRPO)~\cite{deepseekmath/grpo} as our core learning algorithm. Standard GRPO typically normalizes advantages over a single dialogue turn or global sequence. However, in an agentic environment, observation tokens generated by the environment must be excluded from loss calculation. Let $\mathcal{O}_q = \{ o_1, o_2, \dots, o_G \}$ be a group of trajectories generated from a user prompt $q$. For each trajectory $o_i$, we compute a sequence-level reward $r_i$. The advantage function is computed based on the distribution of rewards within the group:
\begin{equation}
    \hat{A}_i = \frac{r_i - \mu_r}{\sigma_r},
\end{equation}
where $\mu_r$ and $\sigma_r$ are the mean and standard deviation of the rewards for the $G$ samples, respectively.

In a standard multi-turn agent setting, the expected objective function is typically formulated as summing over all action tokens:
\begin{equation}
    J(\theta) = \frac{1}{G} \sum_{i=1}^G \frac{1}{\sum_{j=0}^n |a_j|}
 \sum_{j=0}^n \sum_{t=T_j}^{T_j+|a_j|} \min \left[ \frac{\pi_\theta(t|s_t)}{\pi_{\theta_{old}}(t|s_t)} \cdot {A}_{i,t},\ \text{clip}\left(\frac{\pi_\theta(t|s_t)}{\pi_{\theta_{old}}(t|s_t)}, 1-\epsilon, 1+\epsilon\right) \cdot {A}_{i,t} \right].
\end{equation}

However, in the context of Interleaved Thinking, a single trajectory comprises multiple ``Think and Tool-call'' cycles (episodes) with drastically varying action lengths. Simply summing gradients allows episodes to dominate optimization. To ensure balanced supervision across every reasoning step, we propose \textbf{Step-wise Normalization}. We define the optimization objective on individual \textit{Action Segments} $a_j$ rather than the global token stream. Assuming the $i$-th trajectory contains $n_i$ action steps, and the $j$-th action segment $a_j$ has a length of $|a_j|$, our optimized objective function $J(\theta)$ is formalized as:

\begin{equation}
    J(\theta) = \frac{1}{G} \sum_{i=1}^G \frac{1}{n_i} \sum_{j=1}^{n_i} \frac{1}{|a_j|} \sum_{t \in a_j} \min \left[ \frac{\pi_\theta(t|s_t)}{\pi_{\theta_{old}}(t|s_t)} \cdot \hat{A}_{i}, \ \text{clip}\left(\frac{\pi_\theta(t|s_t)}{\pi_{\theta_{old}}(t|s_t)}, 1-\epsilon, 1+\epsilon\right) \cdot \hat{A}_{i} \right].
\end{equation}

This formulation introduces a dual-normalization mechanism:
\begin{enumerate}
    \item \textbf{Action-Step Normalization ($\frac{1}{n_i}$):} Weighs each trajectory equally regardless of the number of ``Think and Tool-call'' cycles.
    \item \textbf{Token-Length Normalization ($\frac{1}{|a_j|}$):} Averages loss within each ``Think and Tool-call'' episode.
\end{enumerate}

\subsubsection{Hybrid Reward}

To steer the model toward both syntactic correctness and factual accuracy, we design a hybrid reward function consisting of three components: Outcome Accuracy Reward, Format Reward, and Hallucination Tool-call Penalty.

\textbf{1. Outcome Accuracy Reward ($R_{acc}$):}
This is a sparse reward computed only at termination. Given the complexity of open-ended multimodal QA, regular expressions are insufficient for verification. We employ a Model-based Judge to evaluate the factual consistency between the model output and the ground truth.
\begin{equation}
    R_{acc} = \begin{cases} 
    1.0 & \text{if Judge returns "1" (Correct)}, \\
    0.0 & \text{if Judge returns "0" (Incorrect)} .
    \end{cases}
\end{equation}

\textbf{2. Format Reward ($R_{fmt}$):}
We implement a strict regex-based parser to enforce schema adherence. This includes:
\begin{itemize}
    \item \textbf{Structural Integrity:} Verifying that tags such as \texttt{<think>}, \texttt{<tool\_call>}, and \texttt{<answer>} appear in valid pairs and sequences.
    \item \textbf{Residue Penalty:} We strictly forbid "chitchat" outside of valid tags (e.g., outputting "I will now execute..." after a \texttt{<tool\_call>} block). Any non-whitespace character outside tags incurs a penalty, as we observed that such residues often lead to output collapse during training.
\end{itemize}
\begin{equation}
    R_{fmt} = \begin{cases} 
    0.5 & \text{if strictly follows schema}, \\
    -0.5 - 0.01 \times \text{len(residue)} & \text{if format error or residue detected}.
    \end{cases}
\end{equation}

\textbf{3. Hallucination Tool-call Penalty ($R_{halluc}$):}
During experiments, \textbf{\textit{we observed a tendency for models to generate consecutive \texttt{<tool\_call>} blocks without waiting for the environment feedback \texttt{<tool\_response>}}}, effectively hallucinating execution results. To suppress this, we penalize the discrepancy between the number of model calls ($N_{call}$) and actual environmental responses ($N_{resp}$):
\begin{equation}
    R_{halluc} = \min(0, (N_{resp} - N_{call}) \times 0.2).
\end{equation}
This mechanism enforces a strict "Turn-taking" protocol, ensuring that only tool calls actually processed by the environment are considered valid behaviors.

The final reward is calculated as:
\begin{equation}
    R_{total} = R_{acc} + \lambda_{fmt} \cdot R_{fmt} + \lambda_{halluc} \cdot R_{halluc}.
\end{equation}
In this paper, we set $\lambda_{fmt}=0.1$ and $\lambda_{halluc}=0.05$.

\subsection{Tool Platform Construction}

\subsubsection{Tool Functions}
In this section, we present the comprehensive multimodal toolkit within MindWatcher, comprising the following five tools:

\textbf{Region Cropping/Zooming:} This tool encompasses diverse image processing operations designed to externalize visual reasoning and highlight critical regions to guide attention. It includes an image grounding tool for localizing and cropping target areas based on input boxes, thereby facilitating the 'thinking with images' reasoning paradigm.

\textbf{Object Grounding and Visual Search:} This tool accepts image interest regions and search categories, subsequently retrieving the corresponding knowledge from a large-scale local image retrieval database (described in Sec~\ref{sec: MRL}). By adaptively localizing query-relevant regions, our tool performs precise and targeted regional searches, effectively addressing complex visual search challenges.

\textbf{External Text Retrieval:} This tool leverages the search engines for information retrieval. It accepts textual queries as input and returns the top-10 ranked results, each comprising a title and an abstract.

\textbf{Webpage Content Extraction:} Taking a URL as input, this tool employs Jina~\cite{jina} to retrieve the webpage content. The agent can read its full content, the content within the window it provides, or use an AI assistant to generate a structured summary based on the specific goal it provides. 

\textbf{Local Code Interpreter:} This tool executes Python code within a sandbox environment isolated from external resources (e.g., files and the internet). It returns the execution results and supports the invocation of various Python libraries for diverse data computation tasks.

\subsubsection{Local Multi-modal Retrieval Library}
\label{sec: MRL}

Conventional image search methods leverage massive Internet data. However, directly acquired Internet resources in fine-grained specialized domains contain erroneous knowledge. Additionally, the high cost of external visual search API calls can significantly increase training expenses during large-scale training.
To alleviate the above issues, we construct a large-scale, high-quality image retrieval database. Based on the general taxonomy of world knowledge, we built our local search database through the following procedure: (1) We established knowledge entries that span eight major categories: Person, Car, Plant, Animal, Logo, Landmark, Fruit \& Vegetable, and Dish. (2) Collect images corresponding to these knowledge entries from both Internet sources and professional museum databases. (3) We employ domain experts to conduct large-scale comparative filtering and knowledge categorization. Through rigorous identification and curation by domain experts, we ensure that the precision of our visual search image database exceeds 99\%. 

Ultimately, our constructed specialized image retrieval database, MindWatcher Multi-modal Retrieval Database\,(MWRD), encompasses eight major categories of knowledge images and associated information, covering a total of 50k retrieval entities. Each retrieval unit contains 3-10 high-quality images, amounting to over 300k images. To accommodate temporally dynamic data, we perform regular maintenance and updates on this image retrieval database.
 
\section{Training Data and MWE-Bench}
The RL training data of MindWatcher includes both online and offline environment training data. The online environment refers to real interactions with internet environments.
In this paper, the online training data comprises three distinct sources: two types of data constructed based on automated pipelines, and data collected from open-source datasets.

\subsection{Training Data Constructed from Private Images}
\label{sec: data_private_image}

To enable MindWatcher agents to master multimodal tools proficiently, we constructed a cross-modal question-answering (QA) dataset with progressively increasing difficulty. Unlike purely textual tasks, this task requires agents to jointly invoke visual perception and external search tools to solve problems.
To ensure data robustness and training efficiency, we designed a Multimodal Knowledge-Augmented Pipeline comprising three core stages: source knowledge anchoring and generation, rigorous QA quality validation, and difficulty grading based on tool invocation.

\subsubsection{Phase 1: Source Knowledge Annotation and Initial Generation}
We first utilize a high-quality private multimodal database as seed data to construct a foundational multimodal dataset. To achieve deep alignment between visual signals and textual knowledge, we designed a generation mechanism comprising the following steps:

\textbf{Fine-Grained Visual-Knowledge Mapping:} We developed an integrated data processing pipeline combining “object localization” and “fine-grained retrieval.” This automated pipeline extracts bounding boxes and corresponding retrieval labels from source images, establishing precise image-text mappings.

\textbf{Knowledge Graph Augmentation:} Based on extracted visual labels, we utilize web search to construct dynamic knowledge graphs, acquiring relevant background knowledge and factual information. This external knowledge is then leveraged to generate initial question-answer pairs, ensuring questions rely not only on images but also integrate external world knowledge.

\subsubsection{Phase 2: Timeliness and Uniqueness Verification}
The accuracy of reward signals is critical in reinforcement learning training. We found that directly generated QA data often faces two major challenges, which may lead to misjudgments in reward models:
\begin{enumerate}
    \item \textbf{Temporal Stability:} Search engine environments are dynamically changing. If a time gap exists between data production and actual training, updates to search results may cause answer drift.
    \item \textbf{Answer Uniqueness and Non-openness:} Open-ended questions often have non-unique solutions. Even if an agent executes the correct search path, its generated answer may contain only partial correct information or be overly broad, making it difficult for the reward model to evaluate.
    In response to these limitations, we implemented a two-stage human-in-the-loop verification pipeline. This rigorous review ensures the final high-quality multimodal dataset maintains temporal consistency, with each question possessing a unique, unambiguous ground truth.
\end{enumerate}
\subsubsection{Phase 3: Difficulty Grading Based on Tool Invocation}
Curriculum learning is an effective strategy for training agents, hinging on reasonable difficulty stratification. However, traditional difficulty assessments based on human subjective perception are often biased.

In tool-integrated scenarios, search engines can instantly resolve memory-based problems deemed “difficult” by humans, creating a disconnect between perceived difficulty and the actual challenges faced by agents. To achieve more precise difficulty screening, we designed a Tool-Invocation Screening Engine. This engine abandons subjective judgments, instead defining sample difficulty through quantitative analysis of the “number of tool invocation rounds” required to solve problems and the “complexity of multi-tool combinations.” This approach constructs training data that truly aligns with the agent learning curve.

\subsection{Training Data Constructed from Open-sourced News}
\label{sec3.2}
Constructing a reliable reward signal for RL in open-ended web interactions is notoriously difficult due to the noisy nature of internet content. General web corpora often abound with subjective commentary, unverifiable rumors, and ambiguous "clickbait" titles, which can lead the reward model to provide incorrect optimization signals. Furthermore, factual information in niche domains is often buried in low-traffic sub-pages that are difficult for generic search engines to index instantly, causing agents to fail even when their reasoning path is correct.

To mitigate these challenges, we selected \textbf{Sports News} as the seed domain for our automated pipeline. Sports data possesses unique characteristics ideal for training TIR agents:

\begin{enumerate}
    \item \textbf{Objective Verifiability:} Unlike social news, sports events have definitive outcomes (scores, winners, rankings) that constitute a unique ground truth.
    \item \textbf{Resistance to Ambiguity:} Statistical facts in sports are less susceptible to the semantic pollution of opinions or fake news.
    \item \textbf{Multimodal Richness:} Match reports are intrinsically multimodal, requiring the alignment of textual statistics with visual evidence (player jerseys, scoreboards, action shots).
\end{enumerate}

We developed a robust Temporal-Aware Multimodal QA Pipeline to harvest and process this data, consisting of three sequential stages: Ingestion, Semantic Auditing, and Constraint-Aware Generation.

\subsubsection{Domain-Specific Ingestion and Filtering}
We deployed a focused crawler targeting authoritative sports portals to ensure information reliability. The raw stream captures article metadata, textual bodies, and associated image sets. A preliminary heuristic filter is applied to discard low-quality samples, retaining only articles with non-empty bodies and at least one relevant image. This creates a raw repository of event-centric multimedia documents.

\subsubsection{Phase 1: LLM-Based Semantic Auditing}
Quality control is paramount for RL training. We introduce a "Data Auditor" agent (powered by a strong LLM) to perform a feasibility check before generation. The auditor evaluates raw news based on a strict \textit{Factuality Protocol}:

\textbf{Retention Criteria:} The content must describe a completed event with a clear timeline (e.g., match results, completed transactions). The text must provide key information (entities, actions) visually corresponding to the images.

\textbf{Rejection Criteria:} Purely subjective content, such as rumors, predictions of future games, gossip, or vague summaries without verifiable details, is discarded.

This phase filters out approximately 40\% of the raw feed, ensuring that the downstream generation model operates only on solid factual ground.

\subsubsection{Phase 2: Constraint-Aware QA Generation}
\label{sec3.2.3}
The surviving samples are processed by a "Question Generator" agent. To prevent the model from learning shortcuts or hallucinating, we designed a Constraint-Aware Prompting Strategy that enforces strict rules on the generated QA pairs:

\textbf{1. Temporal Anchoring:} A critical challenge in time-sensitive QA is "Data Rot"—a question like "Who won the game yesterday?" becomes invalid over time. Our pipeline forces the generator to explicitly resolve relative time expressions into absolute timestamps (e.g., converting "this season" to "the 2025 season") based on the publish time of the article. This ensures the question remains valid and unique indefinitely.

\textbf{2. Visual-Textual Dependency:} Questions are engineered to require information integration from both modalities. For instance, instead of explicitly naming a player, the question might refer to "the player in the No. 8 jersey on the right," compelling the agent to first identify the visual entity and then search for its identity using external knowledge.

\textbf{3. De-referencing Context:} To simulate real-world user queries, we strictly prohibit meta-references such as "According to the article." The agent receives only the standalone question and the image, forcing it to use search tools to retrieve the knowledge originally contained in the source article (which is hidden from the agent during training).

\subsection{Open-sourced Training Data and Offline Training Data}
Supplementing the autonomously constructed online training data detailed previously, we curated a focused collection of open-source datasets. These are strategically utilized to bolster the proficiency of MindWatcher in text-only search tasks and code-augmented mathematical reasoning. 

Furthermore, distinct from the three aforementioned data categories designed for real-world environment training (manual, online-automatic, and open-source), MindWatcher incorporates a specialized offline training method for TIR. To facilitate this, we developed an automated pipeline to construct a substantial corpus of high-quality, multimodal QA pairs with stratified difficulty levels. 

\subsection{MWE-Benchmark}
The MWE-Bench covers six primary categories: Car, Animal, Plant, Person, Landmark, and Sports. While these categories align with those in our automated data construction pipeline, we deliberately adopted a distinct construction methodology for the benchmark to ensure its integrity and prevent performance inflation caused by data-domain overlap. 

Specifically, for data derived from private images, we utilized knowledge entries from our internal database that were strictly excluded from the training set. The construction process followed a multi-stage pipeline: we first expanded our knowledge base by collecting auxiliary web-based information to enrich the context. For each category, we then applied category-specific constraints and employed closed-source models to perform "uniqueness deconstruction"—extracting core factual statements that uniquely identify an entity. These statements formed the basis for constructing initial single-turn QA pairs, which were subsequently synthesized into more complex and challenging multi-step reasoning tasks. Finally, all generated samples underwent a two-tier verification process involving both automated model-based filtering and manual expert review to ensure quality and temporal accuracy.

For sports category data, based on the data construction method outlined in Section~\ref{sec3.2}, we merged text and image corpora belonging to the same entity or event across news data, which are from entirely non-overlapping time points. We then employed a powerful LLM to extract atomic facts from all corpora. Subsequently, we constructed QA pairs with complex queries based on these atomic facts. Finally, data cleaning and filtering were performed following a process similar to that described in Section~\ref{sec3.2.3}.

Following the aforementioned methodology, we have successfully constructed MWE-Bench. The dataset encompasses six categories: 373 car-related instances, 351 animal-related instances, 397 plant-related instances, 63 person-related instances, 90 landmark-related instances, and 142 sports-related instances.

\newpage
\section{Experiment}

\subsection{Application Details}
The training data utilized in this study are segmented across online and offline environments. In the online training environment, we collected VQA data consisting of 1,639 samples based on private images and 2,949 samples derived from public news sources. The open-source domain data, primarily extracted from established benchmarks such as WebSailor~\cite{li2025websailor}, Tool-Star~\cite{tool-star}, and SimpleDeepSearcher~\cite{sun2025simpledeepsearcher}, totaled 5,000 samples. Furthermore, we leveraged approximately 20,000 samples within the offline RL training environment.

The RL process employed a curriculum learning strategy guided by data difficulty. Training was conducted on the Qwen2.5-VL-32B~\cite{bai2025qwen2} model for one epoch. Our training framework features a synchronized rollout mechanism coupled with an asynchronous tool invocation logic\,(Details shown in the Appendix~\ref{secb.2}). Specifically, within each step of the interleaved CoT trajectory, the presence of the \texttt{<im\_end>} token triggers an immediate check for a \texttt{<tool\_call>} token. If a \texttt{<tool\_call>} is detected, it is dispatched instantly. Reward computation also utilizes an asynchronous model invocation method.

We used the fully trained MindWatcher-32B model to distill its multimodal reasoning and tool-use capabilities into smaller, cost-effective models. This process involved collecting an initial, diverse set of base datasets, including the VLAA SFT dataset~\cite{vlaa} (126K samples), the text-only WebWalker silver dataset~\cite{wu2025webwalker} (15K samples), and a self-built multimodal RAG QA dataset (30K samples). The MindWatcher-32B "teacher" model was then employed to roll out and generate 1--3 corresponding TIR trajectories for each sample. After a straightforward filtering process, the final distillation dataset comprised 124K samples, split into 100K multimodal and 24K pure text samples. By using Qwen3-VL-2B~\cite{bai2025qwen3vltechnicalreport}, Qwen2.5-VL-3B, and Qwen3-VL-4B as base models and training them for one epoch on the distillation dataset, we successfully produced the smaller distilled MindWatcher-2B, MindWatcher-3B, and MindWatcher-4B models, respectively.

To comprehensively validate the performance of MindWatcher, in addition to the MWE-Bench, we conducted comparative testing against the model performance on several other open-source benchmarks, including MMSearch\,(subset)~\cite{jiang2024mmsearch}, SimpleVQA\,(subset)~\cite{cheng2025simplevqa}, and WabWalkerQA~\cite{wu2025webwalker}. All tests were conducted with a sampling temperature of $0.7$ and a top-$p$ setting of $0.95$. The primary evaluation metric utilized was $pass@1$, with correctness assessed by employing the LLM-as-Judges methodology.

\subsection{Main Results}
\begin{table}[]
\setlength{\tabcolsep}{8pt}
\renewcommand{\arraystretch}{1.25}
\centering
\caption{Results on the MindWatcher-Evaluation Benchmark.}
\resizebox{\linewidth}{!}{%
\begin{tabular}{lccccccc}
\toprule[1.25pt]
\multicolumn{1}{c}{}                                        & \multicolumn{6}{c}{\textbf{MindWatcher-Evaluation Benchmark}}        \\ \cmidrule{2-7} 
\multicolumn{1}{c}{\multirow{-2}{*}{\textbf{Method}}}              & Car & Animal & Plant & Person & Landmark & Sport & \multirow{-2}{*}{\textbf{Avg.}} \\ \midrule
\multicolumn{8}{c}{\textit{Direct inference}}                                                                \\ \midrule
\multicolumn{1}{l}{GPT-4o mini}                             & 14.48 & 26.21  & 24.43 & 9.52   & 25.56    & 4.23  & 19.63   \\
\multicolumn{1}{l}{GPT-4o}                                  & 24.13 & 38.46  & 29.22 & 12.7   & 38.89    & 6.34  & 27.75   \\
\multicolumn{1}{l}{Qwen2.5-VL-7B}                           & 7.24  & 31.62  & 30.48 & 7.94   & 17.78    & 2.82  & 20.06   \\
\multicolumn{1}{l}{Qwen2.5-VL-32B}                          & 18.5  & 33.62  & 33.25 & 9.52   & 33.33    & 8.45  & 25.92   \\
\multicolumn{1}{l}{Qwen3-VL 8B  Thinking}                   & 10.46 & 22.79  & 23.93 & 15.87  & 17.78    & 16.90 & 18.65   \\
\multicolumn{1}{l}{Qwen3-VL 30B-A3B Thinking}               & 17.69 & 25.64  & 25.69 & 19.05  & 20       & 15.50 & 21.89   \\
\multicolumn{1}{l}{Qwen3-VL 32B Thinking}                   & 14.21 & 28.21  & 26.45 & 20.63  & 21.11    & 21.83 & 22.60   \\
\multicolumn{1}{l}{Openai o4-mini}                          & 21.98 & 35.61  & 31.49 & 11.11  & 47.78    & 6.34  & 27.61   \\
\multicolumn{1}{l}{Gemini 2.5 Pro}                           & 40.48 & 49.57  & 48.11 & 28.57  & 46.67    & 14.09 & 42.09   \\
\multicolumn{1}{l}{Gemini 2.5 Flash}                         & 35.12 & 45.14  & 43.58 & 22.22  & 40       & 17.60 & 37.96   \\
\multicolumn{1}{l}{GPT-5 mini}                              & 22.25 & 48.15  & 42.07 & 4.76   & 44.44    & 13.38 & 33.97   \\ \midrule
\multicolumn{8}{c}{\textit{ReAct/Agent}}                                                                     \\ \midrule
\multicolumn{1}{l}{Gemini 2.5 Flash}                         & \underline{68.17} & 71.76  & 77.26 & 57.14  & 46.51    & 37.32 & 66.65   \\
\multicolumn{1}{l}{GPT-5 mini}                              & 61.93 & 68.66  & 81.61 & 44.44  & \textbf{57.78}    & \textbf{80.28} & \underline{69.91}   \\
\multicolumn{1}{l}{Doubao-Seed-1.6-vision}                   & 57.64 & 57.26  & 61.46 & 65.08  & 38.89    & 39.39 & 57.91   \\
\multicolumn{1}{l}{Qwen2.5-VL-32B}                          & 51.74 & 54.13  & 58.19 & 50.79  & 41.11    & 31.69 & 51.41   \\
\multicolumn{1}{l}{Qwen3-VL 32B Thinking}                   & 58.98 & 75.5   & 77.83 & 69.84  & \underline{48.89}    & \underline{46.48} & 66.95   \\
\multicolumn{1}{l}{WebWatcher-7B}                            & 32.71 & 45.58  & 49.62 & 30.16  & 36.67    & 21.57 &   39.66    \\
\multicolumn{1}{l}{WebWatcher-32B}                           & 38.87 & 53.85  & 67.51 & 52.38  & 42.22    & 32.88 &  50.93  \\
\rowcolor[HTML]{EFEFEF} 
\multicolumn{1}{l}{\cellcolor[HTML]{EFEFEF}MindWatcher-2B}  & 50.13 & 80.34  & 86.4  & 65.08  & 44.44    & 16.90 & 64.76   \\
\rowcolor[HTML]{EFEFEF} 
\multicolumn{1}{l}{\cellcolor[HTML]{EFEFEF}MindWatcher-3B}  & 49.87 & 80.06  & 85.89 & 63.49  & 44.44    & 17.61 & 64.48   \\
\rowcolor[HTML]{EFEFEF} 
\multicolumn{1}{l}{\cellcolor[HTML]{EFEFEF}MindWatcher-4B}  & 56.03 & \underline{84.62}  & \underline{87.66}  & \underline{68.25}  & 41.11    & 36.62  & 69.63  \\
\rowcolor[HTML]{EFEFEF} 
\multicolumn{1}{l}{\cellcolor[HTML]{EFEFEF}MindWatcher-32B} & \textbf{71.31} & \textbf{86.04}  & \textbf{88.92} & \textbf{77.78}  & 47.78    & \underline{46.48} & \textbf{75.35}   \\
\bottomrule[1.25pt]
\end{tabular}
}
\footnotesize
\raggedright
\vspace{2pt}
$^{*}$Best results are in bold and the suboptimal results are in underline. \\
\label{table1}
\vspace{-0.8em}
\end{table}

\begin{table}[b]
\renewcommand{\arraystretch}{1.25}
\centering
\caption{Results on the Open-sourced Benchmarks.}
\resizebox{\linewidth}{!}{%
\begin{tabular}{lccccccc}
\toprule[1.25pt]
                           &                                     & \multicolumn{3}{c}{\textbf{SimpleVQA}}            & \multicolumn{3}{c}{\textbf{WebWalkerQA}}           \\ 
                           \cmidrule(r){3-5} \cmidrule{6-8}
\multirow{-2}{*}{\textbf{Method}} & \multirow{-2}{*}{\textbf{MMSearch}} & Chinese & English & \textbf{Avg.} & Chinese & English & \textbf{Avg.} \\ \midrule
Qwen2.5-VL-32B             & 51.13                               & 21.61            & 21.86            & 21.99        & 42.51            & 17.55            & 26.62        \\
Qwen3-VL 32B thinking      & 57.01                               & 20.78            & 34.85            & 28.91        & 57.09            & 24.71            & 36.47        \\
Gemini 2.5 Flash            & 57.92                               & 30.03            & \underline{39.96}            & 35.93        & 48.18            & 22.17            & 31.62        \\
GPT-5 mini                 & 57.47                               & \underline{32.96}            & 32.9             & 33.29        & \textbf{65.99}            & \textbf{37.41}            & \textbf{47.79}        \\
Doubao-Seed-1.6-vision      & \underline{58.37}                               & \textbf{39.06}            & \textbf{46.1}             & \textbf{43.44}        & \underline{63.56}            & \underline{28.87}            & \underline{41.47}        \\
WebWatcher-7B              & 49.68                               & 21.33            & 26.41            & 24.41        & 40.89            & 19.63            & 27.35        \\
WebWatcher-32B             & 56.45                               & 21.61            & 37.01            & 30.49        & 39.68            & 19.63            & 26.91        \\
\rowcolor[HTML]{EFEFEF} 
MindWatcher-32B            & \textbf{58.82}                               & 22.16            & 39.83            & 32.32        & 50.2             & 23.79            & 33.38        \\ 
\bottomrule[1.25pt]
\end{tabular}
}
\footnotesize
\raggedright
\vspace{2pt}
$^{*}$Best results are in bold and the suboptimal results are in underline. \\
\label{table2}
\end{table}
Tabel~\ref{table1} presents the detailed performance of different backbones on the MWE-Bench under both direct inference and React/Agent inference modes.

\textbf{Disparity in Parametric Knowledge.} 
Under the Direct Inference mode, we observe that the freshness of a model knowledge cutoff does not linearly correlate with its benchmark performance. Despite being the most recent release, the Qwen3-VL series achieves an average score of only 22.60. In contrast, Gemini 2.5 Pro—notwithstanding an older knowledge cutoff—attains a SOTA zero-shot score of 42.09. This discrepancy underscores a critical reality: when internal parameters alone are insufficient for handling long-tail or specialized world knowledge, the integration of external reasoning tools is necessary.

\textbf{Tool-Augmented Performance Leap.}
Transitioning to the ReAct/Agent paradigm catalyzes a significant performance surge for models previously limited by their internal knowledge. For instance, the score of Qwen3-VL 32B 
nearly triples when equipped with tool-use capabilities. Similarly, GPT-5 mini exhibits a remarkable explosion in performance within the Sports domain, soaring from 13.38 to 80.28 upon gaining tool access.

\textbf{MindWatcher Dominance.} 
MindWatcher-32B achieves overall SOTA performance on MWE-Bench with a global score of 75.35, outperforming prominent closed-source commercial models such as Gemini2.5 Flash and GPT-5 mini. Notably, MindWatcher achieves the highest accuracy across four specific domains: Vehicle, Animal, Plant, and Person. Furthermore, the distilled variants, including MindWatcher-2B, 3B, and 4B, demonstrate performance comparable to the Qwen3-VL 32B baseline. This empirically demonstrates that robust tool-call capabilities can effectively mitigate the knowledge gaps typically present in small-parameter models.

Table~\ref{table2} presents the comparative performance of MindWatcher-32B against other models in identical environments across two filtered multimodal subsets (MMSearch and SimpleVQA) and one pure-text benchmark (WebWalkerQA). MindWatcher continues to deliver SOTA results on MMSearch among all open- and closed-source models evaluated. On the SimpleVQA subset, MindWatcher performance surpasses the next-generation Qwen3-VL-32B base model. Importantly, on the pure-text WebWalkerQA benchmark, MindWatcher remains highly competitive. Compared to its base model, Qwen2.5-VL-32B,  results indicate that our continuous multimodal agentic RL training has successfully enhanced agent capabilities without compromising its foundational text-based reasoning.
\begin{wrapfigure}{r}{0.57\textwidth} 
    \centering
    
    \begin{subfigure}{\linewidth}
        \centering
        \includegraphics[trim={1.5cm 9cm 3cm 1.5cm}, clip, width=\textwidth]{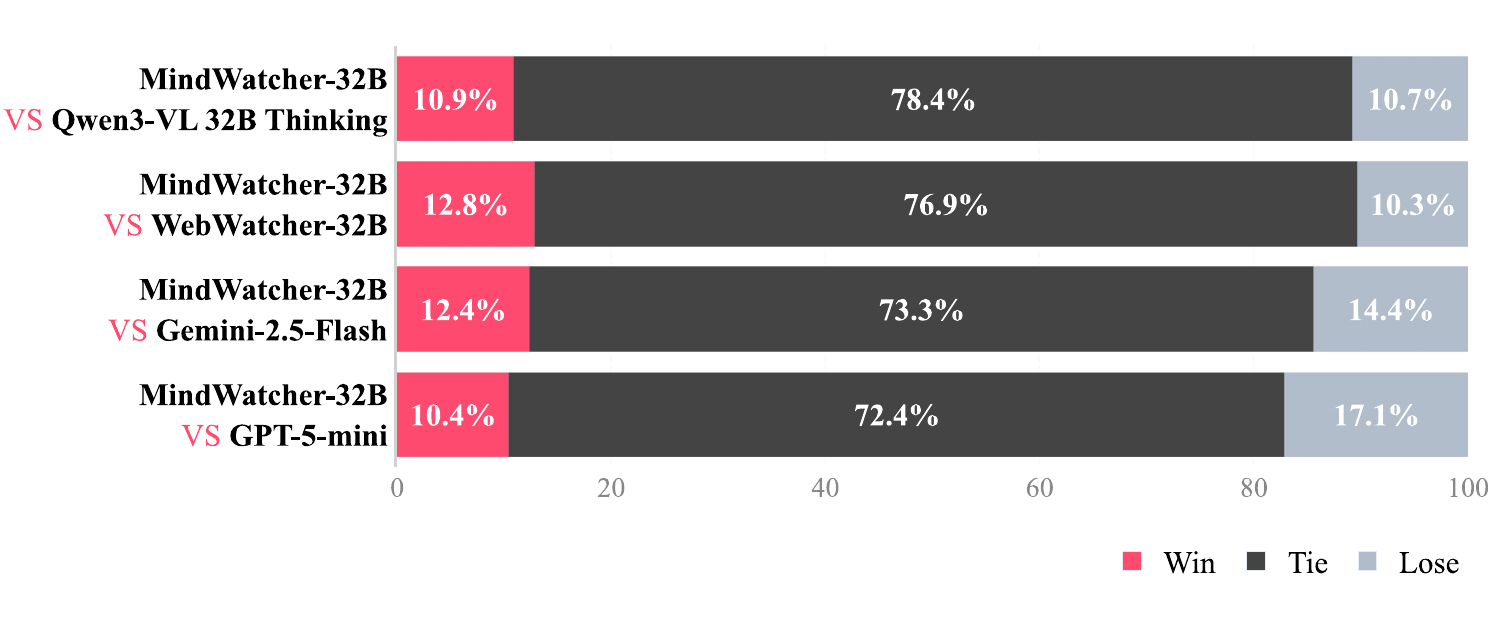}
        \caption{Open-sourced Benchmark.}
        \label{fig:result_a}
    \end{subfigure}
    
    \vspace{0.5em} 
    
    \begin{subfigure}{\linewidth}
        \centering
        \includegraphics[trim={1.5cm 9cm 3cm 1.5cm}, clip, width=\textwidth]{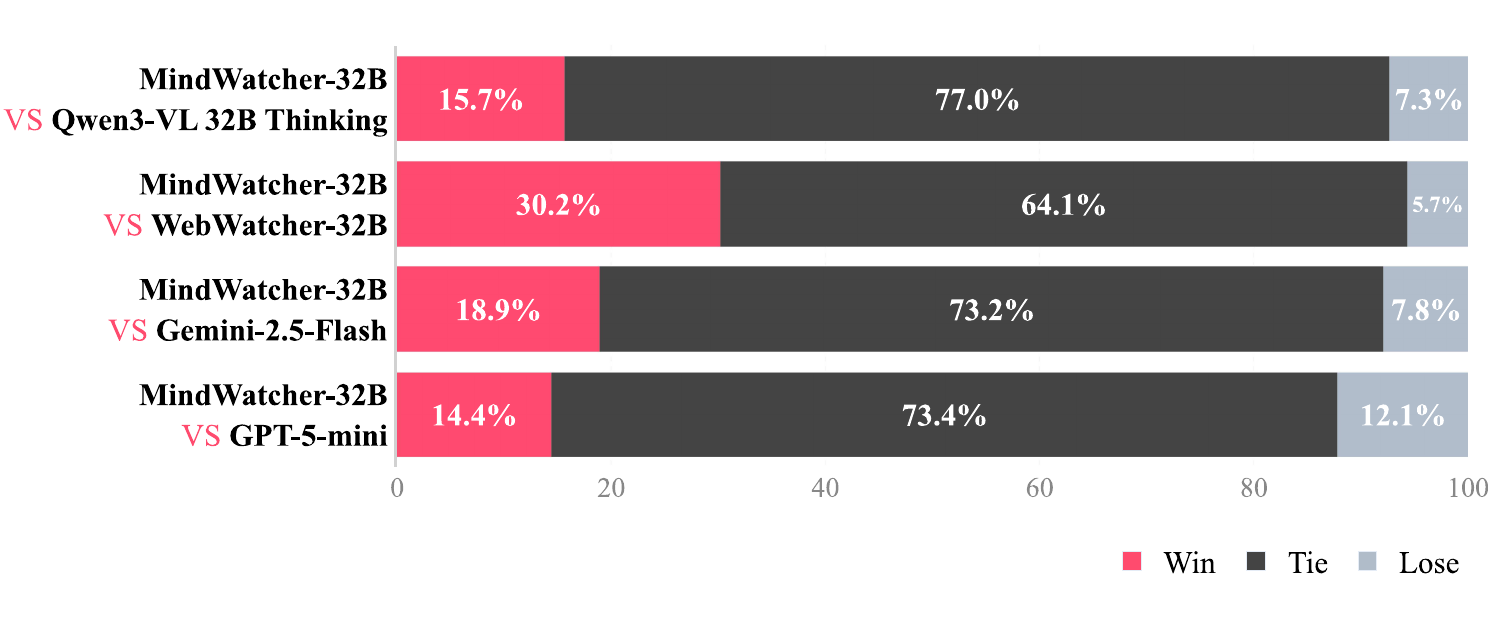}
        \caption{MWE-Benchmark.}
        \label{fig:result_b}
    \end{subfigure}
    
    \vspace{-2pt} 
    \caption{Benchmark Performance Comparison.}
    \label{fig:win-tie-loss}
    \vspace{-3pt} 
\end{wrapfigure}

Figure~\ref{fig:win-tie-loss} presents the win-tie-loss analysis comparing MindWatcher-32B against four representative models: Qwen3-VL 32B Thinking, WebWatcher-32B, Gemini 2.5 Flash, and GPT-5 mini, across both public open-source benchmarks and our MWE-Bench. The results indicate that MindWatcher-32B consistently outperforms its parameter equivalent 32B counterparts in both evaluation settings. Notably, on the MWE-Bench, MindWatcher-32B demonstrates superior performance even when compared to SOTA closed-source models, specifically Gemini 2.5 Flash and GPT-5 mini.

Table~\ref{table3} further details the performance gains achieved by the three distilled small-scale models relative to their respective foundation models on the MWE-Bench. Among these, MindWatcher-3B (derived from Qwen2.5-VL-3B-Instruct) exhibits the most significant improvement, with its proficiency score surging from 24.93 to 64.48. This substantial leap underscores the effectiveness of our distilled training approach in empowering small-scale models with robust agentic capabilities.

\begin{table}[]
\setlength{\tabcolsep}{8pt}
\renewcommand{\arraystretch}{1.2}
\caption{Comparison Results of the Distilled Models and their Base Models.}
\resizebox{\linewidth}{!}{%
\begin{tabular}{cccccccc}
\toprule[1.25pt]
                                                           & \multicolumn{6}{c}{\textbf{MindWatcher-Evaluation Benchmark}}         \\ \cmidrule(r){2-7} 
\multirow{-2}{*}{\textbf{Method}}                                 & Car   & Animal & Plant & Person & Landmark & Sport & \multirow{-2}{*}{\textbf{Avg.}} \\ 
\midrule  
Qwen3-VL 2B Thinking                                       & 39.95 & 70.37  & 80.1  & 50.79  & 24.44    & 29.51 & 51.41   \\
\rowcolor[HTML]{ECF4FF} 
\multicolumn{1}{l}{\cellcolor[HTML]{ECF4FF}MindWatcher-2B} & 50.13 & 80.34  & 86.4  & 65.08  & 44.44    & 16.90 & 64.76   \\
Qwen2.5-VL 3B Instruct                                     & 17.43 & 35.04  & 37.78 & 6.3    & 12.22    & 0.00  & 24.93   \\
\rowcolor[HTML]{ECF4FF} 
\multicolumn{1}{l}{\cellcolor[HTML]{ECF4FF}MindWatcher-3B} & 49.87 & 80.06  & 85.89 & 63.49  & 44.44    & 17.61 & 64.48   \\
Qwen3-VL 4B Thinking                                       & 56.3  & 81.2   & 80.86 & 74.6   & 38.89    & 32.79 & 66.53   \\
\rowcolor[HTML]{ECF4FF} 
\multicolumn{1}{l}{\cellcolor[HTML]{ECF4FF}MindWatcher-4B} & 56.03 & 84.62  & 87.66 & 68.25  & 41.11    & 39.51 & 69.63   \\ 
\bottomrule[1.25pt]
\end{tabular}
}
\label{table3}
\vspace{-0.5em}
\end{table}

\subsection{Analysis}

\subsubsection{The Impact of the Tool Capacity}
During experiments, we find that the proficiency of the integrated tools is a pivotal determinant of an agent's final performance. This is particularly evident in external retrieval tasks, where the indexing and recall mechanisms of different search engines lead to highly heterogeneous outcomes for identical queries. Beyond direct downstream performance, we observed that the choice of search engine during RL training induces distinct tool-call behavioral adaptations and search patterns within the model.

To quantify this impact, we conducted experiments using sports-related datasets, subdivided into two domains (Football and Basketball) and two languages (Chinese and English). We evaluated the agent performance using three search engines—Sogou, Bing, and Quark—under a retrieval-only setting. The results, summarized in Table~\ref{table4}, demonstrate a substantial performance variance that frequently overshadows the variations attributed to algorithmic optimizations or foundation model scales.
Specifically, in the most extreme case (football queries written by Chinese), the Quark search engine outperformed Sogou by a staggering 42.86\%. However, these findings do not point to a universally "superior" search engine; rather, we found that the effectiveness of a search engine is highly volatile and contingent upon the specific domain and language of the query. This volatility highlights that the "capacity" of an agent is deeply coupled with its environment, suggesting that benchmark evaluations must account for tool-induced variance to ensure a fair assessment of a model's intrinsic reasoning abilities.
\begin{table}[t]
\renewcommand{\arraystretch}{1.15}
\centering
\caption{Results on different search engines.}
\resizebox{0.8\linewidth}{!}{
\begin{tabular}{lccccc}
\toprule[1.25pt]
\multicolumn{1}{c}{\multirow{2}{*}{\textbf{Search Engine}}} & \multicolumn{2}{c}{\textbf{Basketball}} & \multicolumn{2}{c}{\textbf{Football}} & \multirow{2}{*}{\textbf{Avg.}} \\ \cmidrule(r){2-3} \cmidrule(r){4-5}
\multicolumn{1}{c}{}                               & English            & Chinese             & English            & Chinese            &                      \\ \midrule
Sogou Search                                        & 2.53           & 15.19          & 3.57          & 12.5          & 8.51                 \\ 
Bing search                                         & \underline{13.92}          & \underline{20.25}          & \underline{8.93}          & \underline{23.21}         & \underline{16.66}                \\
Quark Search                                        & \textbf{20.25}          & \textbf{39.24}          & \textbf{28.57}         & \textbf{55.36}         & \textbf{34.81}                \\

\bottomrule[1.25pt]
\end{tabular}}
\label{table4}

\footnotesize
\vspace{2pt}
$^{*}$Best results are in bold and the suboptimal results are in underline. \\
\end{table}

\subsubsection{Genetic Inheritance in Agentic RL}
\label{sec4.3.2}
\begin{figure}[b]
    \centering
    \begin{subfigure}{0.49\textwidth}
        \centering
        \includegraphics[trim={1.5cm 12.5cm 1.5cm 2.5cm}, clip, width=\textwidth]{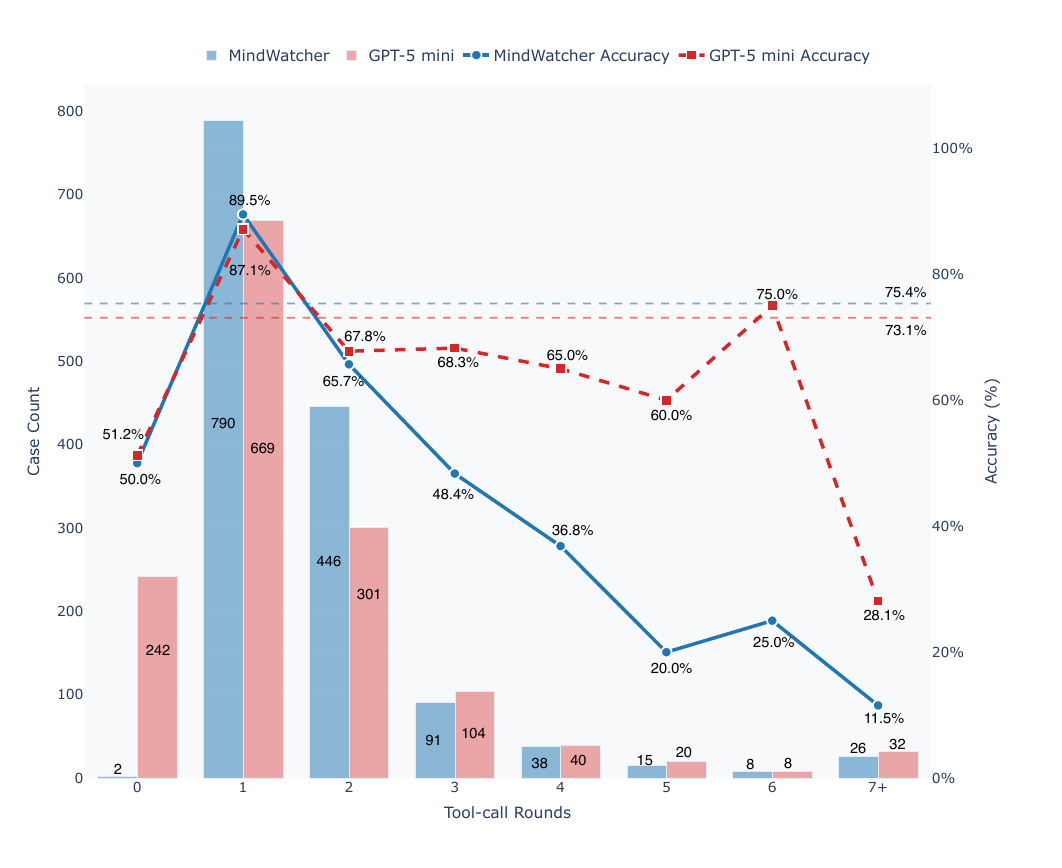}
        \caption{MindWatcher vs GPT-5 mini.}
        \label{fig:sub-b}
    \end{subfigure}
    \begin{subfigure}{0.49\textwidth}
        \centering
        \includegraphics[trim={1.5cm 12.5cm 1.5cm 2.5cm}, clip, width=\textwidth]{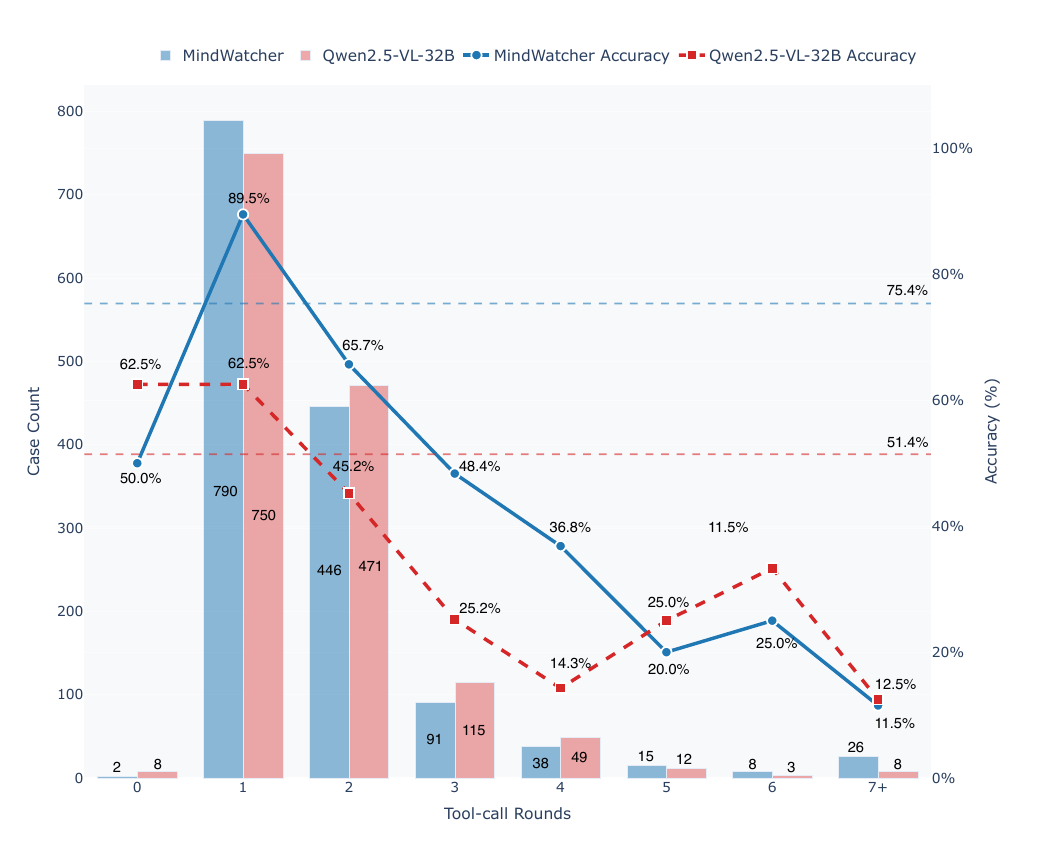}
        \caption{MindWatcher vs Qwen2.5-VL-32B.}
        \label{fig:sub-a}
    \end{subfigure}
    \hfill 
    
    \caption{Comparison of Tool-use Behavior and Performance Distribution.}
    \label{fig4}
\end{figure}

We conduct a granular analysis of the relationship between tool-calling frequency and model accuracy. Specifically, we compare the behaviors and performance of MindWatcher against its own foundation model, Qwen2.5-VL-32B, and GPT-5 mini on the WME-Bench. The visualization of these results is presented in Figure~\ref{fig4}.

\textbf{Disparity in Decision Trigger Boundaries:} A significant divergence is observed in the decision-making boundaries regarding tool invocation. As illustrated, GPT-5 mini opts to reason without any tool-calls (Round 0) in 
nearly one-sixth of the samples, yet achieves a mere 51.2\% accuracy. This suggests a manifest "blind self-confidence" in GPT-5 mini; by relying on internal parameters for tasks requiring external knowledge, it forfeits substantial scores at the onset, leading to a lower overall performance compared to MindWatcher. Interestingly, when the number of tool-calls exceeds one, GPT-5 mini exhibits remarkable robustness in long-chain reasoning, with its accuracy showing negligible decay from Round 2 to Round 6. 

\begin{case}[label={case1}]{The Impact of Model World Knowledge on Agent Performance.}
    \begin{center}
        \includegraphics[width=0.3\linewidth]{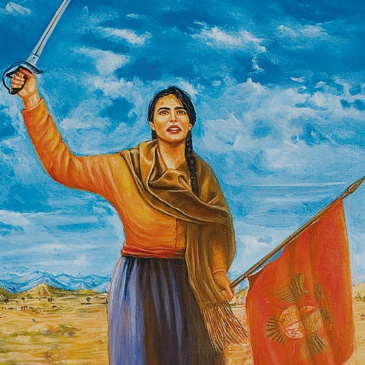}   \\
    \end{center}

    \textbf{Question:} In the image, a significant battle took place during the early 19th century in this northern South American city where a prominent female patriot played a crucial role by leading a company of soldiers and providing essential support to a key military leader's forces. This patriot was also known for her contributions to the independence movement of her country. In the main cathedral of this city, how many women are interred? \\
    \textbf{Ground-truth:} 2 
    \begin{stepTitle}{\textbf{Qwen3-VL 32B:}}
    \textbf{<think>} \\
    Okay, let's try to figure this out. The user is asking about a northern South American city where a significant battle happened in the early 19th century. \dots The flag has a red color with some emblem. Let me think. In South American independence, key figures include Simón Bolívar, and female figures like \textcolor{red}{\textbf{Manuela Sáenz}}. Wait, \textcolor{red}{\textbf{Manuela Sáenz}} was a prominent female patriot in the independence movement of Bolivia or Peru? \dots
    \textbf{<\textbackslash think>} \\
    \dots   
    \end{stepTitle}

    \begin{stepTitle}{\textbf{MindWatcher 32B:}}
        \textbf{<think>} \\
        The question provides enough context to proceed. The woman is described as a prominent female patriot who played a crucial role in a significant battle during the early 19th century in a northern South American city. She led a company of soldiers and provided essential support to a key military leader's forces. She is also known for her contributions to the independence movement of her country.   \\
        Given this information, \textcolor{blue}{\textbf{I can use the `web\_search` tool to search for prominent female patriots in northern South American cities who were involved in significant battles during the early 19th century}}. Once I identify the correct patriot, I can determine the city she is associated with and then search for how many women are interred in the main cathedral of that city.  \\
        \textbf{<\textbackslash think>} \\
        \textbf{<tool\_call>}    \\
        \{
        "name": "web\_search",
        "arguments": \{
            "query": prominent female patriot in northern South America early 19th century,
        \}
        \}  \\
        \textbf{<\textbackslash tool\_call>} \\  
        \dots
    \end{stepTitle}
\end{case}
This phenomenon highlights that for high-capacity models, agentic performance can be bottlenecked by the decision trigger boundary rather than the executive action capability itself. Under autonomous settings, the model potential can be severely constrained by its initial failure to recognize the need for external tools.

\textbf{Performance Shadowing and Genetic Inheritance in Agentic RL:} While MindWatcher, trained via RL, significantly outperforms its foundation model (Qwen2.5-VL-32B), we observe a profound "Genetic Inheritance" in reasoning capacity. This is evidenced by the striking consistency in both accuracy trends and sample distribution across different tool-calling rounds.

As the required number of tool-calls increases, MindWatcher maintains a higher accuracy than Qwen2.5-VL-32B but fails to reverse the downward trend (identical decay slope) inherited from its foundation.
Furthermore, MindWatcher’s self-awareness—manifested in its sample distribution across varying tool-call rounds—shows no significant deviation from Qwen2.5-VL-32B even after extensive RL training.

These observations suggest that \textbf{while agentic RL can substantially refine tool-invocation and reasoning proficiency, it cannot fully breach the performance bottlenecks of the foundation model regarding long-range reasoning and multimodal processing.} The foundation model imposes a fundamental performance constraint on the RL-derived agent; agentic RL serves as a strategy optimizer but remains fundamentally coupled with the base model’s capabilities. We term this phenomenon the \textbf{"Genetic Constraint" of the foundation model in agentic RL scenarios}.
In Appendix~\ref{seca.3}, we conducted further investigations into genetic inheritance in the agentic SFT scenario.

\subsubsection{The Impact of Model World Knowledge on Agent Performance}

Beyond numerical analysis, we conduct a case-based qualitative study to investigate how the world knowledge inherent in different foundation models affects downstream task performance. We observe that when the provided tools are insufficient for a "low-knowledge" model to resolve a query, the model internal world knowledge becomes the decisive factor for downstream benchmark metrics.

Case Study~\ref{case1} presents a visual comparison between MindWatcher (based on Qwen2.5-VL-32B) and the next-generation Qwen3-VL 32B Thinking on a specific case. In this example, neither model can correctly answer the question based solely on their internal capabilities. However, once provided with an external text-retrieval tool, a significant performance gap emerges: Qwen3-VL possesses the internal world knowledge to recognize the name of the person\,(Manuela Sáenz) in the artwork. This prior knowledge allows it to formulate precise search queries using the text-search tool, leading to a successful resolution. 
Conversely, MindWatcher (based on Qwen2.5-VL) lacks any prior information regarding this specific artwork. Without a starting point for inquiry and lacking auxiliary tools to bridge this knowledge gap, the model is unable to perform any viable reasoning or retrieval.

This case study demonstrates that in identical tool environments, performance metrics may not exclusively reflect TIR capabilities of a model. Current benchmarks contain a significant number of tasks that implicitly rely on the "long-tail" knowledge of a model to catalyze the tool-use process. This coupling introduces substantial challenges in isolating and accurately evaluating the intrinsic TIR capacity of a model, as the benchmark results become confounded by the uneven distribution of world knowledge across different foundation models.
For a given benchmark, when a vast number of queries cannot be adequately addressed by the provided tools, the evaluation of the agent functional capabilities essentially regresses into a test of the model's internal world knowledge.

\section{Related Work}
\subsection{TIR Agent}

The landscape of TIR agents has witnessed a meteoric evolution over the past six months. By empowering models to autonomously select and invoke tools, the capability boundaries of LLMs—particularly those with smaller parameter counts—have been significantly expanded. However, a stark contrast persists: while contemporary LLMs exhibit reasoning capabilities comparable to human experts (often likened to "PhD-level" cognition), their \textit{action competence}—specifically, the precision and robustness of tool invocation—remains at a nascent, almost "elementary" stage.

OpenAI o3~\cite{openai_o3}, as the first TIR agent deployed to a global user base, demonstrated astonishing proficiency. By actively manipulating images, performing complex calculations via code execution, and navigating file systems, o3 illuminated the vast potential of TIR agents to the research community. This paradigm shift catalyzed a surge of open-source initiatives inspired by o3, ranging from specialized code and search agents to DeepResearch systems.
rStar2-Agent~\cite{shang2025rstar2} leverages code execution as a verifier and solver to bolster mathematical reasoning. DeepEyes~\cite{zheng2025deepeyes} introduces active visual tools, such as "image zoom-in," probing the ability of multimodal agents to resolve fine-grained visual details through iterative manipulation. The Qwen DeepResearch~\cite{li2025webweaver,geng2025webwatcher, tongyideep} team has also made pivotal contributions to the open-source ecosystem by systematically diagnosing and addressing the multi-dimensional challenges inherent in long-horizon research tasks.
Despite these strides, the chasm between an agent's "thinking" and "acting" remains substantial. Critical challenges such as dynamic tool context management, long-term historical memory maintenance, and the attainment of training-free tool invocation capabilities represent significant hurdles that the field must address in the near future.

\subsection{Training Method of TIR Agent}
Unlike traditional LLM training, training TIR agents presents distinct challenges due to the necessity of interacting with external environments—specifically, executing tool calls and interpreting heterogeneous feedback during generation. Beyond mere planning, agents must learn to act adaptively within dynamic information contexts.
Several works focus on the offline training stages of continual pre-training and SFT. \cite{scaling-agent} proposes an agent-specific continued pre-training method designed to endow base models with native action capabilities, thereby effectively supporting subsequent fine-tuning. WebDancer~\cite{wu2025webdancer} and WebSailor~\cite{li2025websailor} concentrate on methodologies for constructing high-quality TIR trajectories; while they incorporate elements of RL, they predominantly rely on SFT to shape agent behavior.
The transition to online RL, which requires interaction with real-world environments, precipitates a steep rise in training complexity. Recent research has tailored algorithms specifically for this regime. ARPO~\cite{arpo} introduces an entropy balancing mechanism to prevent the training collapse often observed during TIR agent RL. LLDS~\cite{grpo_collapse} investigates the "lazy likelihood displacement" problem in agent RL, introducing likelihood preservation regularization to avert systemic stagnation in training. In summary, given the high cost of constructing premium trajectory data and the inherent instability of online RL, the path toward an optimal training strategy for TIR agents remains long and arduous.

\section{Conclusion}
In this paper, we introduced MindWatcher, a high-performance TIR agent. We developed robust visual question answering training data construction pipelines and proposed a specialized training methodology distinct from prior works. To empower MindWatcher, we established a comprehensive yet cost-effective multimodal toolbox and introduced MWE-Bench for rigorous performance evaluation. Our experiments demonstrate that MindWatcher, through its superior tool-invocation capabilities, can match or even exceed the performance of significantly larger or updated models. Beyond empirical results, this study reveals several critical experimental findings discovered during our development of the TIR agent. We hope our work provides unique insights and contributes to the future advancement of tool-augmented intelligence.

\newpage

\begin{ack}
We extend our gratitude to Tongyi Qwen for their outstanding contributions to open-source LLMs; and we thank all colleagues at LiAuto Base Model for their support of the MindWatcher project.
\end{ack}

\section*{Author List}
\paragraph{Core Contributors (Equal contribution)} Jiawei Chen \;\;\; Xintian Shen \;\;\; Lihao Zheng \;\;\; Zhenwei Shao

\paragraph{Contributors} Handong Cui \;\;\; Chaoqun Du \;\;\; Li Gong \;\;\; Feng Gu \;\;\; Xuefeng Hao \;\;\; Wei He \;\;\; Jiabang He \;\;\; Yi Hu \;\;\; Bin Huang \;\;\; Shanshan Li \;\;\; Qizhen Li \;\;\; Jing Luo \;\;\; Zide Liu \;\;\; Xiaobo Liu \;\;\; Ning Mao \;\;\; Lifu Mu \;\;\; Xuhao Pan \;\;\; Zhiheng Qu \;\;\; Chang Ren \;\;\; Xudong Rao \;\;\; Haoyi Sun \;\;\; Qian Wang \;\;\; Shuai Wang \;\;\; Zhichao Wang \;\;\; Wei Wang \;\;\; Lian Wen \;\;\; Jiqing Zhan \;\;\; Hongfu Yang \;\;\; Sheng Yang \;\;\; Jiajun Yang \;\;\; Pengfei Yu \;\;\; Hongyuan Zhang \;\;\; Bin Zhang \;\;\; Chunpeng Zhou \;\;\; Zheng Zhou \;\;\; Shucheng Zhou \;\;\; Shuo Xie \;\;\; Yun Zhu

\paragraph{Technique Leaders} Hao Ma \;\;\; Tao Wei

\paragraph{Supervisors} Pan Zhou \;\;\; Wei Chen

\newpage
\appendix

\section{Technical Appendices and Supplementary Material}
\subsection{Open-sourced Benchmark}
In this work, rather than utilizing the full set or a naive random subset of open-source benchmarks, we implemented a rigorous data filtration pipeline. This decision stems from the observation that many existing benchmarks suffer from significant limitations, such as information lag due to insufficient temporal constraints. Furthermore, models released after benchmark publication may exhibit inflated performance due to inadvertent data leakage. 

To address these issues, we established the Qwen3-VL 32B Thinking as a baseline for direct inference on the original benchmarks. All samples correctly answered by the model through direct inference were discarded. For the remaining samples, we conducted a meticulous manual review to filter out ambiguous questions or those with expired time-sensitivity. This process yielded a high-quality subset of open-source benchmarks, which we subsequently used to evaluate reasoning and tool-integrated capabilities under the ReAct/Agent paradigm.

Among these, the MMSearch subset contains 221 samples, while the simplevqa subset comprises 823 samples, including 361 Chinese samples and 462 English samples.
\subsection{Infrastructure}
\label{secb.2}
\begin{figure}[h]
    \centering
    \includegraphics[trim={5cm 15cm 5cm 0cm}, clip, width=\textwidth]{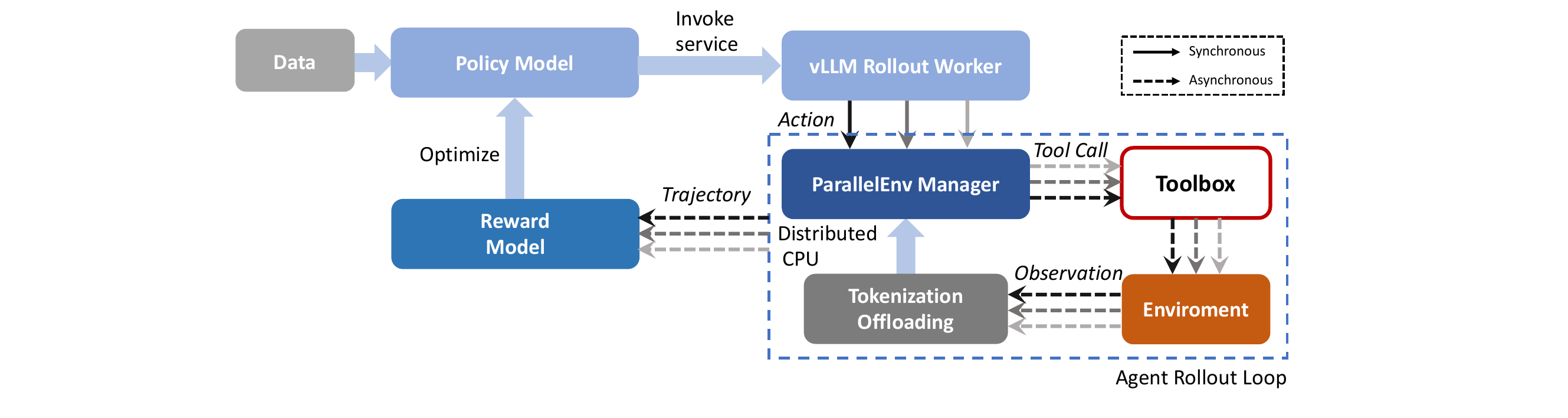}
    \caption{Step-wise Synchronous Sampling Framework of MindWatcher.}
    \label{fig:infra}
\end{figure}

To facilitate efficient agentic reinforcement learning, we developed a step-wise synchronous sampling framework based on the Verl~\cite{verl} to coordinate interactions between the agent and external environments as shown in Figure~\ref{fig:infra}. In each rollout iteration, the vLLM engine~\cite{vllm} performs parallel batch inference to generate actions, followed by a synchronized barrier where the environment collects feedback. This design ensures trajectory consistency across massive batches while simplifying state management for the on-policy training process.

Empirical observations during training revealed that the primary bottleneck is not the trajectory generation itself, as the latency gap between synchronous and asynchronous sampling remains marginal. Instead, the dominant time expenditure arises from tool-calling latency. To mitigate this, we integrated an asynchronous tool invocation layer within the synchronous loop. By leveraging asyncio mechanisms and semaphore-based concurrency control, heterogeneous tools are dispatched and executed in parallel while strictly adhering to API QPS constraints. Furthermore, we implemented Tokenization Offloading, which offloads the computationally intensive task of processing environment observations' tokenization from the master node to distributed CPU workers. Additionally, the LLM-as-a-Judge reward model is invoked immediately upon the completion of each trajectory to minimize evaluation overhead. 
This hybrid architecture—synchronous in step control but asynchronous in tool execution—maximizes hardware utilization and significantly reduces the actual rollout time.

\subsection{Genetic Inheritance in Agentic SFT}
\label{seca.3}

\begin{figure}[hp]
    \centering
    \begin{minipage}{1.0\textwidth}
        \centering
        \includegraphics[trim={1.0cm 8.0cm 1.0cm 2.0cm}, clip, width=\textwidth]{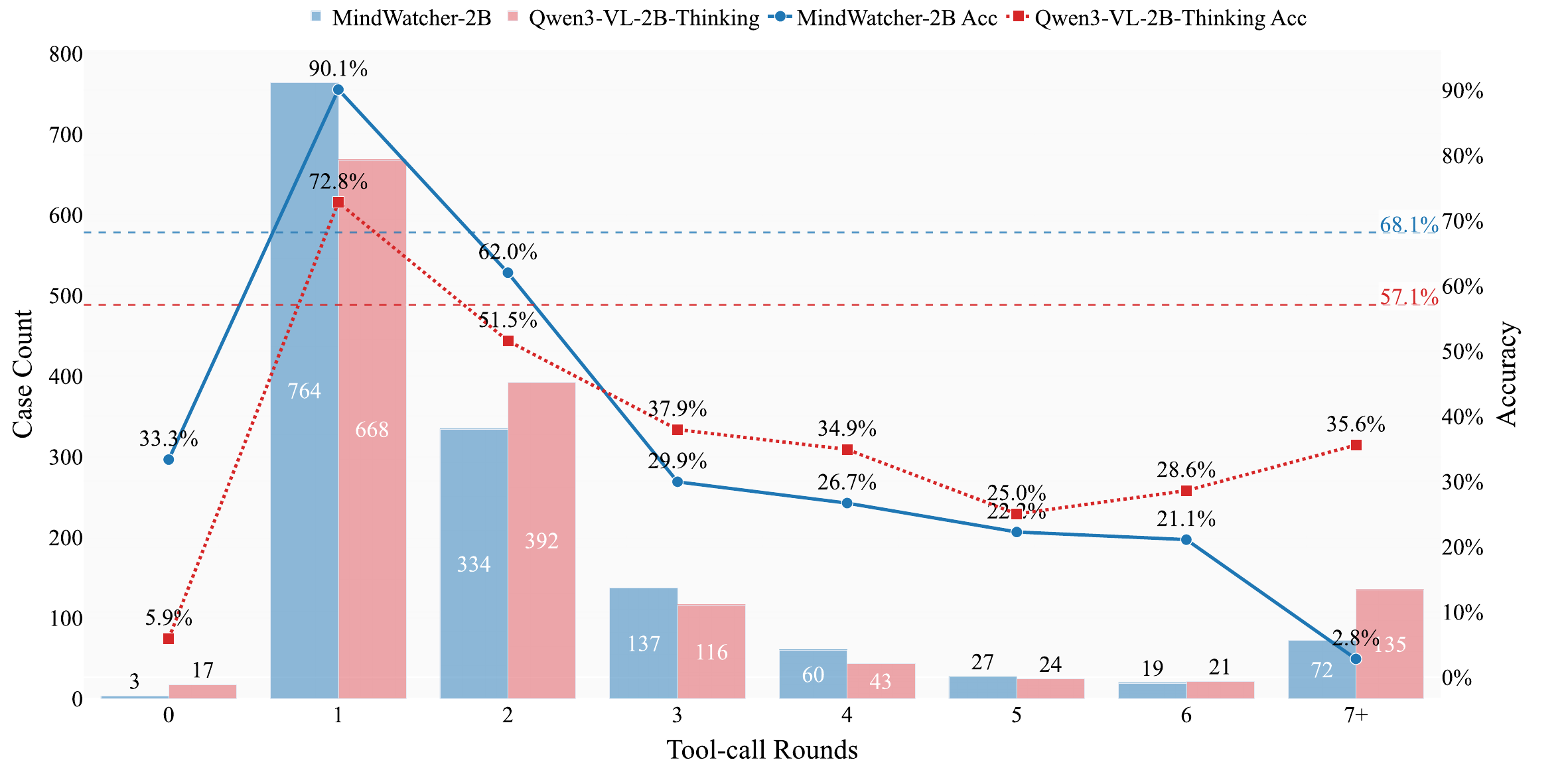}
        \caption{MindWatcher-2B vs Qwen3-VL 2B Thinking.}
        \label{fig5}
    \end{minipage}
    
\hfill 

    \begin{minipage}{1.0\textwidth}
        \centering
        \includegraphics[trim={1.0cm 8.0cm 1.0cm 2.0cm}, clip, width=\textwidth]{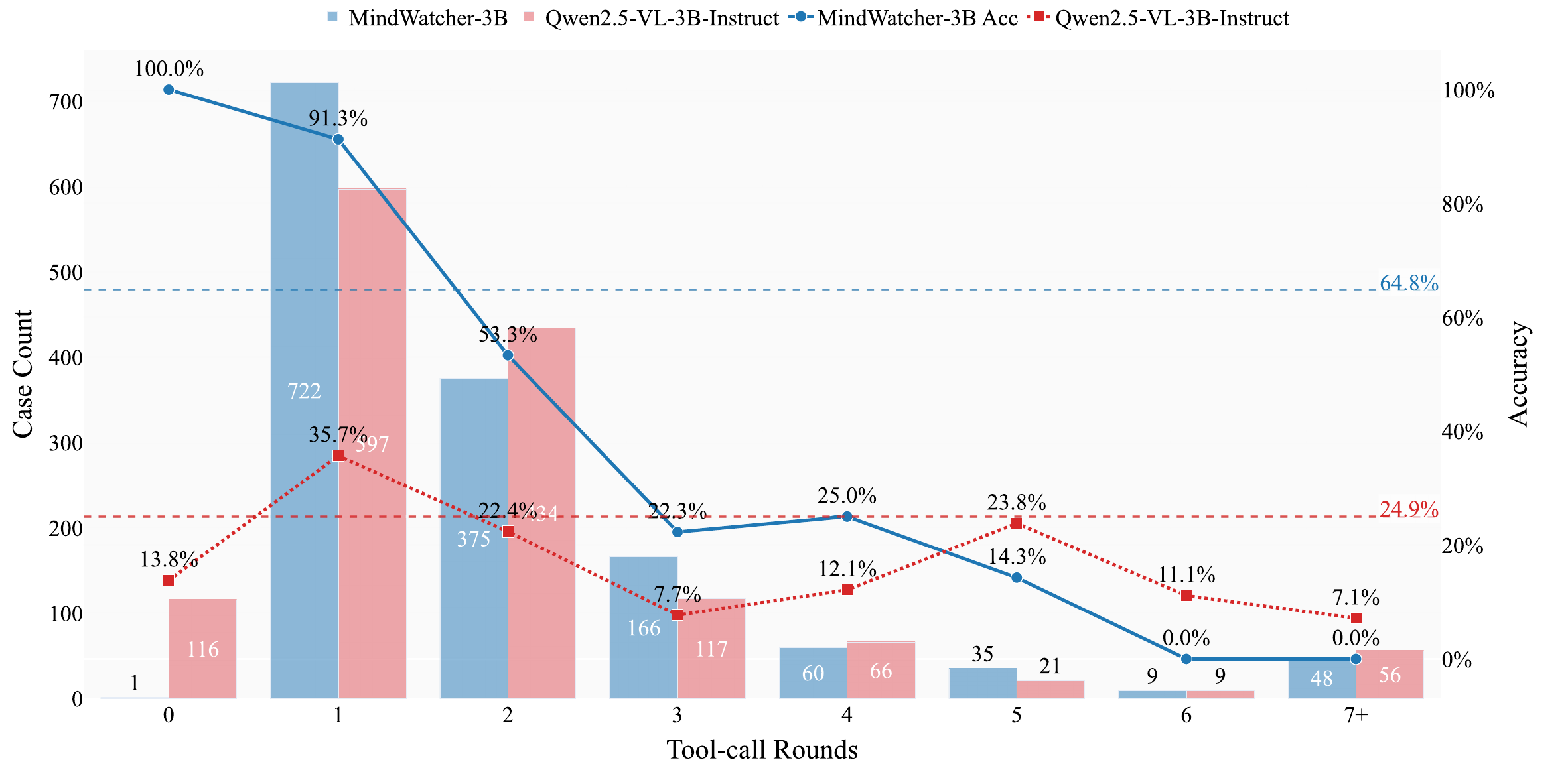}
        \caption{MindWatcher-3B vs Qwen2.5-VL-3B.}
        \label{fig6}
    \end{minipage}

\hfill 
    \begin{minipage}{1.0\textwidth}
        \centering
        \includegraphics[trim={1.0cm 8.0cm 1.0cm 2.0cm}, clip, width=\textwidth]{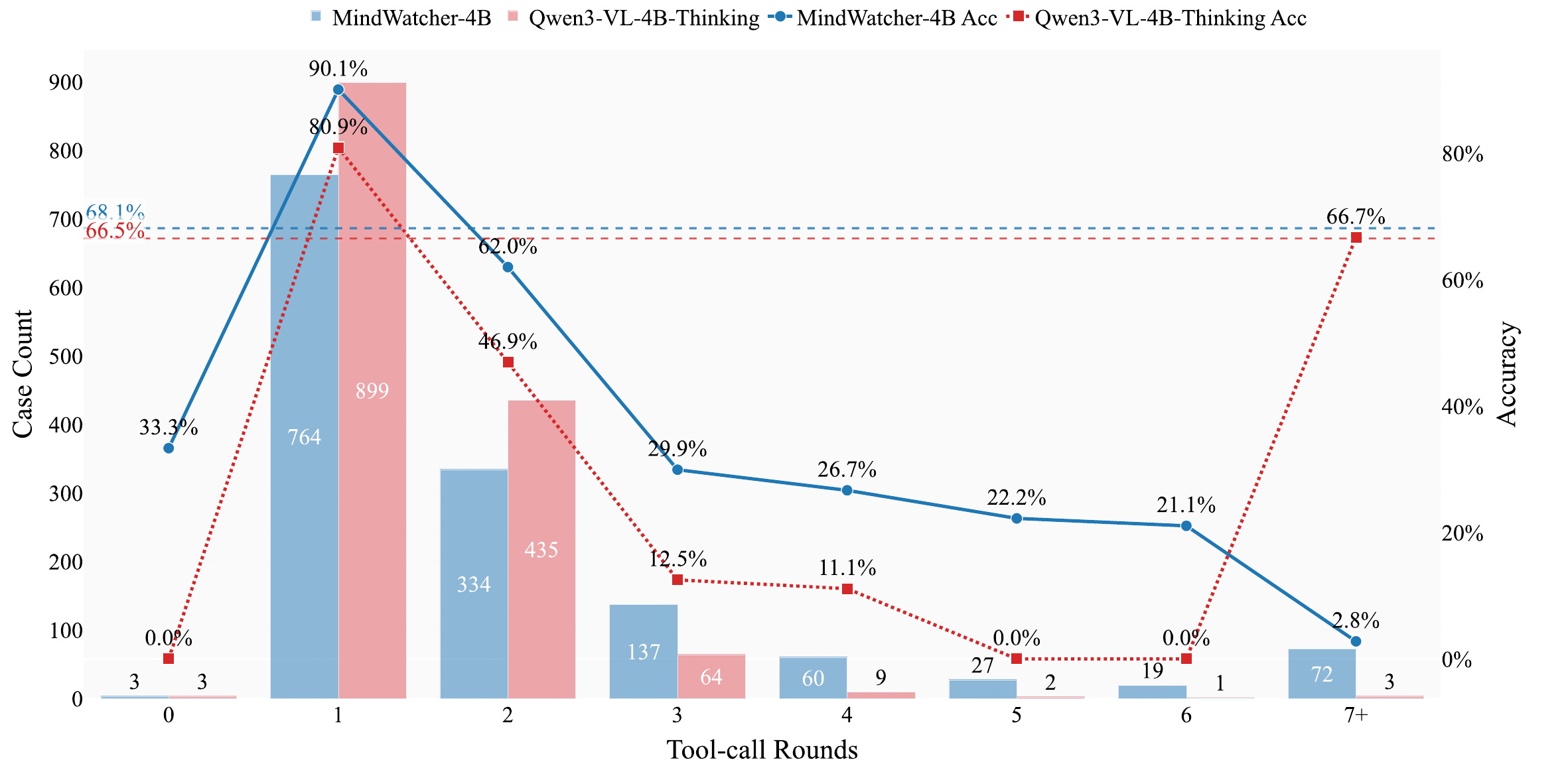}
        \caption{MindWatcher-4B vs Qwen3-VL 4B Thinking.}
        \label{fig7}
    \end{minipage}
\end{figure}

Building upon our analysis of genetic inheritance in the agentic RL paradigm (Section~\ref{sec4.3.2}), we extend our investigation to the agentic SFT scenario using three distilled small-scale agent models. Figures \ref{fig5}, \ref{fig6}, and \ref{fig7} illustrate the tool-use behavior and performance distributions of MindWatcher-2B, 3B, and 4B alongside their respective foundations: Qwen3-VL-2B Thinking, Qwen2.5-VL-3B, and Qwen3-VL-4B Thinking.

Our observations indicate that, unlike the RL scenario, SFT-tuned models do not exhibit a consistent or predictable trend in tool-calling frequency relative to their base models. The decision trigger boundary in the SFT paradigm appears significantly less robust.
For instance, after agentic SFT, the Qwen2.5-VL-3B model showed a dramatic shift, with Round 0 cases (no tool use) plummeting from 116 to just 1.
Across the three distilled models, the distribution of tool-call rounds fluctuates inconsistently before and after SFT, lacking the stable behavioral alignment observed in the RL-tuned MindWatcher-32B.

Despite the behavioral volatility, the accuracy trends across different tool-call rounds reveal a phenomenon strikingly similar to that of agentic RL. As the complexity of the task increases (i.e., more tool-call rounds), both the SFT-tuned agents and their base models exhibit a synchronized downward trend in accuracy.

This reinforces the existence of "genetic inheritance" within the SFT paradigm: supervised fine-tuning is inherently limited by the base model capabilities in long-range reasoning and multimodal processing.
Like RL, SFT serves as a method for policy alignment but fails to break through the fundamental cognitive "ceiling" established by the foundation model.

A key distinction arises in the "elegance" of the performance curves. In the agentic RL scenario, the performance of the agent and the base model decay at nearly identical slopes, showing a highly structured coupling. In contrast, the performance curves in the SFT are less congruent; while they share the same downward trajectory, the lack of a perfectly parallel slope suggests that SFT introduces more noise or less systematic optimization into the model’s reasoning-tool-use integration compared to the more rigorous RL process.

\newpage
\subsection{Tool description for MindWatcher}
\begin{toolbox}{Tool: Region Croping/Zooming}
    \textbf{Description:} Zoom into a specific area of **the first input image** based on your provided bounding box. \\
    \textbf{input:} the image and bounding box. \\
    \textbf{output:}a new image.  \\
    \textbf{Arguments:}
    \begin{itemize}
        \item \texttt{bbox}: [x1, y1, x2, y2], \# The bounding box you provided, where (x1, y1) is the top-left corner and (x2, y2) is the bottom-right corner.
    \end{itemize}
\end{toolbox}

\begin{toolbox}{Tool: Object Grounding \& Visual Search}
    \textbf{Description:} Retrieve new similar images and their descriptions based on the provided bounding box area of **the first input image**. \\
    \textbf{input:}image and bounding box.  \\
    \textbf{output:}only the most similar target's type name and the confidence score.  \\
    \textbf{Arguments:}
    \begin{itemize}
        \item \texttt{bbox\_2d:}[x1, y1, x2, y2], \# The bounding box you provided, where (x1, y1) is the top-left corner and (x2, y2) is the bottom-right corner.
        \item \texttt{category:}"the category" \# The category of the image you want to search for, which can only be one of \{plant, animal, car, person, landmark, vegetable, cuisine, logo\}.
        
    \end{itemize}
\end{toolbox}

\begin{toolbox}{Tool: External Text Retrieval}
    \textbf{Description:} Retrieve external text information from the internet based on your provided text query. \\
    \textbf{input:}only text query. \\
    \textbf{output:}text.   \\
    \textbf{Arguments:}
    \begin{itemize}
        \item \texttt{query:} "the content".
    \end{itemize}
\end{toolbox}

\begin{toolbox}{Tool: Webpage Content Retrieval}
    \textbf{Description:} Visit a specified web page URL under 3 modes: you can read its full content, the content within the window you provided, or use an AI assistant to generate a structured summary based on the specific goal you provide.  \\
    \textbf{input:}A JSON object containing three arguments: url, window, goal.\\
    \textbf{output:}A JSON object containing the visited URL and a structured result.   \\
    \textbf{Arguments:}
    \begin{itemize}
        \item \texttt{\text{url}:} "https://example.com/article", \# The webpage url you want to visit.
        \item \texttt{window:} [a, b], \# Select the content you want to read within [a, b] (Optional)
        \item \texttt{goal:} "".  \# What you want to get or find. (Optional)
    \end{itemize}
\end{toolbox}

\begin{toolbox}{Tool: Local Code Interpreter}
    \textbf{Description:} Execute Python code in a sandboxed environment. Useful for calculations, data manipulation, and other programmatic tasks. This tool is isolated and cannot access files or the internet. \\
    \textbf{input:}A string of Python code. The tool call format can be a standard JSON object or a Python dictionary literal. \\
    \textbf{output:}A JSON object containing stdout (from print statements), stderr (error messages), and the final result value.  \\
    \textbf{Arguments:}
    \begin{itemize}
        \item \texttt{code}: "\text{the code such as: a = 5\textbackslash nb = 10\textbackslash nresult = a * b\textbackslash nprint(result)}" \# The Python code to execute.
    \end{itemize}
\end{toolbox}

\newpage
\subsection{Trajectory Display}

In this section, we visualize tool calling trajectory of MindWatcher to highlight the model multimodal chain-of-thought reasoning and interleaved thinking capabilities.

\begin{toolbox}{Case1}
    \begin{center}
        
    \includegraphics[width=0.5\linewidth]{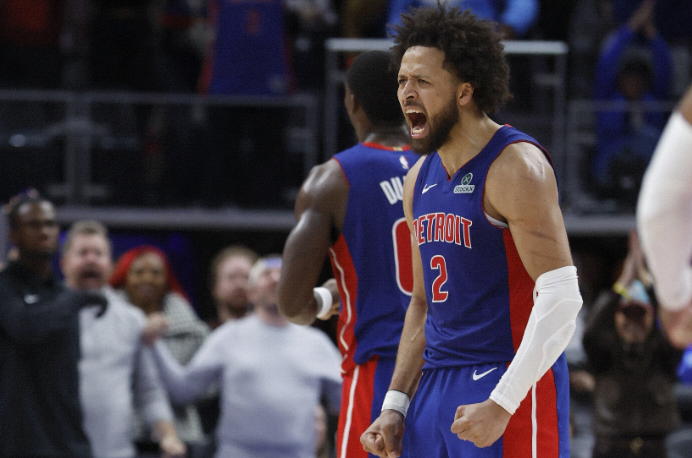}   \\
    \end{center}
    
    \label{fig:placeholder}
    \vspace{1pt}
    \textbf{Question:} 在2025年11月18日，图中这名球员与耐克续签代言后，他成为现役NBA/WNBA与耐克仍有签名鞋合同的第几位球星？截至2025年11月18日，图中这名来自底特律球队的球员的个人耐克签名鞋预计在哪个赛季首次亮相？在2025年11月18日这次续约时，图中球员所属的联赛共有多少位现役球员拥有与耐克的签名鞋合同？以2025年11月18日为准，与图中这名球员所属联赛不同的另一联赛共有多少位现役球员拥有与耐克的签名鞋合同？在2025年11月18日的情况下，图中这名身穿“DETROIT”球衣的球员与耐克续签的代言合同期限是几年？ \\
    \textbf{Ground-truth:}第9位; 2026-27赛季; 6位; 3位; 6年 
    \vspace{0.5pt}
    \begin{stepTitle}{\textbf{Step 1:}}
        \textcolor{liautoblue}{
    \textbf{<think>} \\
    为了回答这个问题，我需要识别出图片中的球员，并获取以下信息：\\
1. 截至2025年11月18日，这位球员与耐克续签代言后，成为现役NBA/WNBA与耐克仍有签名鞋合同的第几位球星。\\
2. 截至2025年11月18日，这位球员的个人耐克签名鞋预计在哪个赛季首次亮相。\\
3. 在2025年11月18日这次续约时，这位球员所属的联赛共有多少位现役球员拥有与耐克的签名鞋合同。\\
4. 截至2025年11月18日，与这位球员所属联赛不同的另一联赛共有多少位现役球员拥有与耐克的签名鞋合同。\\
5. 截至2025年11月18日，这位球员与耐克续签的代言合同期限是几年。\\
首先，我需要识别出图片中的球员。可以通过调用工具来帮助我完成这个任务。\\
    \textbf{<\textbackslash think> \\}}
    \textbf{<tool\_call>}    \\
    \{
    "name": "zoom\_v\_search",
    "arguments": \{
        "bbox\_2d": [1300, 0, 2761, 2209], 
        "category": "person" 
    \}
    \}  \\
    \textbf{<\textbackslash tool\_call>} \\  
    \textbf{<tool\_response>}    \\
    名称：凯德·坎宁安, 检索置信度：0.81   \\
    \textbf{<\textbackslash tool\_response>}
    \end{stepTitle}

    \begin{stepTitle}{\textbf{Step 2:}}
        \textcolor{liautoblue}{
        \textbf{<think>} \\
        通过使用工具，我识别出图片中的球员是凯德·坎宁安（Cade Cunningham）。接下来，我需要查找与他相关的耐克代言信息，包括他在2025年11月18日与耐克续签代言后的排名、签名鞋首次亮相的赛季、续约时现役球员中有多少人拥有与耐克的签名鞋合同、WNBA中拥有耐克签名鞋合同的现役球员数量以及他与耐克续约的合同期限。为了获取这些信息，我将使用网络搜索工具。\\
        \textbf{<\textbackslash think>} \\
        }
        \textbf{<tool\_call>}    \\
            \{
            "name": "web\_search",
            "arguments": \{
                "query": 凯德·坎宁安 2025年11月18日 耐克 续签
            \}
        \}  \\
    \textbf{<\textbackslash tool\_call>} 
 
    \end{stepTitle}
\end{toolbox}
\begin{toolbox}{Case1}

    \textbf{<tool\_response>}    \\
\{ \\
    \hspace*{1em} \texttt{'search\_query'}: '凯德·坎宁安 2025年11月18日 耐克 续签', \\
    \hspace*{1em} \texttt{'search\_result'}: [ \\
    \hspace*{2em} \{ \\
    \hspace*{3em} \texttt{'id'}: 1, \\
    \hspace*{3em} \texttt{'title'}: '下一位超巨！坎宁安与耐克达成6年续约：签名鞋将于下赛季亮相', \\
    \hspace*{3em} \texttt{'content'}: '北京时间11月18日，底特律活塞队的后卫凯德·坎宁安将继续与耐克公司保持合作关系... 根据知名记者Shams的报道，坎宁安已经与耐克签署了一份新的为期六年的代言合同...', \\
    \hspace*{3em} \texttt{'url'}: 'https://m.sohu.com/a/956031720\_121722346/?pvid=000115\_3w\_a', \\
    \hspace*{3em} \texttt{'date'}: '2025-11-19T10:43:00+08:00' \\
    \hspace*{2em} \}, \\
    \hspace*{2em} \dots \\
    \hspace*{2em} \{ \\
    \hspace*{3em} \texttt{'id'}: 8, \\
    \hspace*{3em} \texttt{'title'}: '耐克续约坎宁安，他将成为后詹杜库时代美国本土球员耐克的新门面', \\
    \hspace*{3em} \texttt{'content'}: '耐克把六年合同拍在坎宁安面前... 2027年才能买到的第一双CC战靴，其实今天就已经在生产线排号...', \\
    \hspace*{3em} \texttt{'url'}: 'https://m.toutiao.com/a7574051184151937574/' \\
    \hspace*{2em} \} \\
    \hspace*{1em} ] \\
    \} \\
    \textbf{<\textbackslash tool\_response>} \\
    \begin{stepTitle}{\textbf{Step 3:}}
        \textcolor{liautoblue}{
        \textbf{<think>} \\
        通过网络搜索，我获得了以下信息：\\
        1. 凯德·坎宁安在2025年11月18日与耐克续签了一份为期六年的代言合同。\\
        2. 他的签名鞋预计将在2026-27赛季的下半程正式发布。\\
        3. 他是现役第六位拥有耐克签名鞋的NBA球员，与勒布朗·詹姆斯、凯文·杜兰特、扬尼斯·安特托昆博等超级巨星并列顶尖阵营。\\
        根据这些信息，我可以回答问题：\\
        1. 截至2025年11月18日，凯德·坎宁安与耐克续签代言后，成为现役NBA/WNBA与耐克仍有签名鞋合同的第6位球星。\\
        2. 截至2025年11月18日，凯德·坎宁安的个人耐克签名鞋预计在2026-27赛季首次亮相。\\
        3. 在2025年11月18日这次续约时，凯德·坎宁安所属的联赛（NBA）共有6位现役球员拥有与耐克的签名鞋合同。\\
        4. 关于WNBA中拥有耐克签名鞋合同的现役球员数量，我没有找到具体的信息。为了回答这个问题，我将使用网络搜索工具。\\
        5. 截至2025年11月18日，凯德·坎宁安与耐克续签的代言合同期限是6年。\\
        接下来，我将使用网络搜索工具来查找WNBA中拥有耐克签名鞋合同的现役球员数量。\\
        \textbf{<\textbackslash think>} \\
        }
        \textbf{<tool\_call>}    \\
        \{
        "name": "web\_search",
        "arguments": \{
            "query": WNBA 现役 耐克 签名鞋 合同 数量
        \}
        \}  \\
        \textbf{<\textbackslash tool\_call>} \\  
        \textbf{<tool\_response>}    \\
        \{ \\
        \hspace*{1em} \texttt{'search\_query'}: 'WNBA 现役 耐克 签名鞋 合同 数量', \\
        \hspace*{1em} \texttt{'search\_result'}: [ \\
        \hspace*{2em} \{ \\
        \hspace*{3em} \texttt{'id'}: 1, \\
        \hspace*{3em} \texttt{'title'}: 'Nike签约6位NBA及3位WNBA球员拥有签名鞋', \\
        \hspace*{3em} \texttt{'content'}: '目前拥有Nike签名鞋合同的6位NBA球员和3位WNBA球员。目前拥有Nike签名鞋合同的6位NBA球员和3位WNBA球员。...', \\
        \hspace*{3em} \texttt{'url'}: 'http://cj.sina.cn/articles/view/2432009827/p90f5866302701jglm', \\
        \hspace*{3em} \texttt{'date'}: '2025-11-18T12:00:00+08:00' \\
        \hspace*{2em} \}, \\
        \hspace*{2em} \dots \\
    \end{stepTitle}    
\end{toolbox}
\begin{toolbox}{Case1}

    \hspace*{2em} \{ \\
    \hspace*{3em} \texttt{'id'}: 8, \\
    \hspace*{3em} \texttt{'title'}: '现役拥有Nike签名鞋合同9位球星！NBA6人 WNBA3人！', \\
    \hspace*{3em} \texttt{'content'}: '11月18日讯 据Shams Charania报道，活塞球星康宁汉姆和耐克续签一份为6年的代言合同，并将拥有自己的签名鞋。球员的个人签名鞋预计将26-27赛季首次亮相。康宁汉姆也成为了NBA/WNBA现役球员中 位与耐克还有签名鞋合同的球星...', \\
    \hspace*{3em} \texttt{'url'}: 'http://17449.xdgk.com.cn/dianshiju/38952adc0ec9b8fNw6lg.html' \\
    \hspace*{3em} \texttt{'date'}: '2025-12-20T01:37:34+08:00' \\
    \hspace*{2em} \} \\
    \hspace*{1em} ] \\
    \} \\
    <\textbackslash tool\_response> \\
    \begin{stepTitle}{\textbf{Step 4:}}
    \textcolor{liautoblue}{
    \textbf{<think>} \\
    通过网络搜索，我获得了以下信息：\\
    1. WNBA中拥有耐克签名鞋合同的现役球员数量为3人。\\
    现在我已经有了所有问题的答案：\\
    1. 截至2025年11月18日，凯德·坎宁安与耐克续签代言后，成为现役NBA/WNBA与耐克仍有签名鞋合同的第6位球星。\\
    2. 截至2025年11月18日，凯德·坎宁安的个人耐克签名鞋预计在2026-27赛季首次亮相。\\
    3. 在2025年11月18日这次续约时，凯德·坎宁安所属的联赛（NBA）共有6位现役球员拥有与耐克的签名鞋合同。\\
    4. 截至2025年11月18日，与凯德·坎宁安所属联赛不同的另一联赛（WNBA）共有3位现役球员拥有与耐克的签名鞋合同。\\
    5. 截至2025年11月18日，凯德·坎宁安与耐克续签的代言合同期限是6年。\\
    我将给出最终答案。\\
    \textbf{<\textbackslash think>}} \\
    \end{stepTitle}
\textcolor{textred}{
\textbf{<answer>}    \\
1. 截至2025年11月18日，凯德·坎宁安与耐克续签代言后，成为现役NBA/WNBA与耐克仍有签名鞋合同的第6位球星。\\
2. 截至2025年11月18日，凯德·坎宁安的个人耐克签名鞋预计在2026-27赛季首次亮相。\\
3. 在2025年11月18日这次续约时，凯德·坎宁安所属的联赛（NBA）共有6位现役球员拥有与耐克的签名鞋合同。\\
4. 截至2025年11月18日，与凯德·坎宁安所属联赛不同的另一联赛（WNBA）共有3位现役球员拥有与耐克的签名鞋合同。\\
5. 截至2025年11月18日，凯德·坎宁安与耐克续签的代言合同期限是6年。\\
\textbf{<\textbackslash answer>} \\
}
\end{toolbox}
\begin{toolbox}{Case2}
        \begin{center}
        \includegraphics[width=0.5\linewidth]{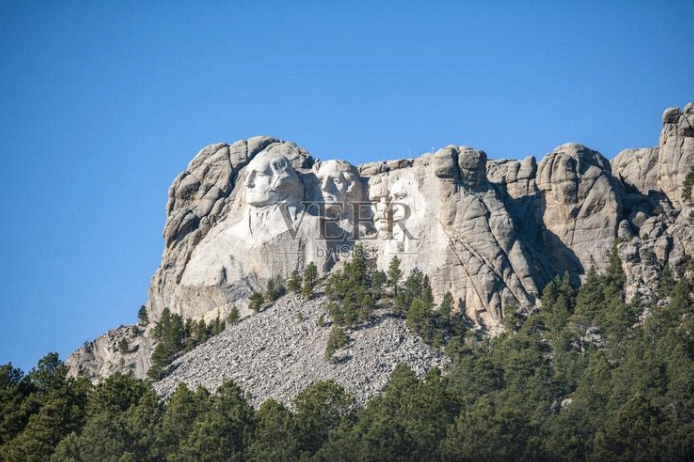}   \\    
        \end{center}
        \label{fig:placeholder}
    \textbf{Question:} 这座山体雕刻上刻画了哪几位美国总统的面孔？其中哪一位人物在独立战争中发挥了关键作用？他哪一年写了信件，这封信授权美国首次正式开展间谍活动？ \\
    \textbf{Ground-truth:}刻画了乔治·华盛顿、托马斯·杰斐逊、西奥多·罗斯福、亚伯拉罕·林肯；乔治·华盛顿在美国独立战争中发挥了关键作用；1777年 
    \begin{stepTitle}{\textbf{Step 1:}}
        \textcolor{liautoblue}{
            \textbf{<think>} \\
            这张图片显示的是美国著名的地标——拉什莫尔山（Mount Rushmore）。根据图片中的特征，我可以识别出这座山体雕刻上刻画了几位美国总统的面孔。为了确认具体是哪几位总统，并找到与独立战争相关的关键人物及其写信授权美国首次正式开展间谍活动的年份，我需要调用工具来获取更多信息。\\
            \textbf{<\textbackslash think>} \\
        }
        \textbf{<tool\_call>}    \\
        \{
        "name": "zoom\_v\_search",
        "arguments": \{
            "bbox\_2d": [186, 145, 637, 356], 
            "category": "landmark" 
        \}
        \}  \\
        \textbf{<\textbackslash tool\_call>} \\  
        \textbf{<tool\_response>}    \\
        名称：拉什莫尔山国家纪念公园, 检索置信度：0.97   \\
        \textbf{<\textbackslash tool\_response>} \\
    \end{stepTitle}
    \vspace{-1em}
    \begin{stepTitle}{\textbf{Step 2:}}
        \textcolor{liautoblue}{
            \textbf{<think>} \\
            根据工具返回的结果，确认这座山体雕刻是拉什莫尔山国家纪念公园。拉什莫尔山上雕刻了四位美国总统的面孔：乔治·华盛顿、托马斯·杰斐逊、西奥多·罗斯福和亚伯拉罕·林肯。现在，我需要找出其中哪一位在独立战争中发挥了关键作用，并确定他写信授权美国首次正式开展间谍活动的年份。为了获取这些详细信息，我将使用web\_search工具来查找相关信息。\\
            \textbf{<\textbackslash think>} \\    
        }
        \textbf{<tool\_call>}    \\
        \{
        "name": "web\_search",
        "arguments": \{
            "query": 拉什莫尔山 四位美国总统 独立战争 关键人物 首次正式开展间谍活动
        \}
        \}  \\
        \textbf{<\textbackslash tool\_call>} \\  
        \textbf{<tool\_response>}    \\
        \{ \\
        \hspace*{1em} \texttt{'search\_query'}: '拉什莫尔山 四位美国总统 独立战争 关键人物 首次正式开展间谍活动', \\
        \hspace*{1em} \texttt{'search\_result'}: [ \\
        \hspace*{2em} \{ \\
        \hspace*{3em} \texttt{'id'}: 1, \\
        \hspace*{3em} \texttt{'title'}: '美国南达科他州拉什莫尔山上的四个人头像是哪四位啊', \\
        \hspace*{3em} \texttt{'content'}: '美国南达科他州拉什莫尔山上的四个人头像是哪四位啊美国南达科他州拉什莫尔山上的四座高60英尺的总统头像，象征着美国建国150年来的历史。这四位总统分别是乔治·华盛顿、托马斯·杰弗逊、西奥多·罗斯福和亚伯拉罕·林肯。华盛顿作为美...', \\
        \hspace*{3em} \texttt{'url'}: 'https://zhidao.baidu.com/question/572951197473818884.html', \\
        \hspace*{3em} \texttt{'date'}: '2024-10-23T00:00:00+08:00' \\
        \hspace*{2em} \}, \\
        \hspace*{2em} \dots \\
        \hspace*{2em} \{ \\
        \hspace*{3em} \texttt{'id'}: 5, \\
        \hspace*{3em} \texttt{'title'}: '独立战争期间美国反间谍第一人', 
    \end{stepTitle}
\end{toolbox}
\begin{toolbox}{Case2}
    \begin{stepTitle}{}

        \hspace*{3em} \texttt{'content'}: '独立战争期间美国反间谍第一人,美国,威廉,杰伊,约翰,华盛顿,反间谍,克劳斯比,独立战争', \\
        \hspace*{3em} \texttt{'url'}: 'https://m.163.com/dy/article/K30AB0GV0514R9P4.html'' \\
        \hspace*{3em} \texttt{'date'}: '2025-06-26T16:33:59+08:00' \\
        \hspace*{2em} \} \\
        \hspace*{1em} ] \\
        \} \\
        \textbf{<\textbackslash tool\_response>} \\
    \end{stepTitle}
    \begin{stepTitle}{\textbf{Step 3:}}
        \textcolor{liautoblue}{
            \textbf{<think>} \\
            通过工具返回的搜索结果，我了解到拉什莫尔山上雕刻的四位美国总统是乔治·华盛顿、托马斯·杰斐逊、西奥多·罗斯福和亚伯拉罕·林肯。为了找到在独立战争中发挥了关键作用的人物以及他写信授权美国首次正式开展间谍活动的年份，我需要进一步查看搜索结果中的相关内容。根据第三个搜索结果，提到独立战争期间美国反间谍第一人，并且涉及威廉·杰伊和约翰·华盛顿等人物。为了获取更多详细信息，我将访问相关网页以获取具体细节。\\
            \textbf{<\textbackslash think>} \\
        }
        \textbf{<tool\_call>}    \\
        \{ 
            "name": "url\_visit", "arguments": \{ "url":"https://m.163.com/dy/article/K30AB0GV0\\514R9P4.html", "goal": "找到在独立战争中发挥了关键作用的人物以及授权年份" \} \} \\
            \textbf{<\textbackslash tool\_call>} \\ 
            \textbf{<tool\_response>}    \\
        \{ \\
            \hspace*{1em} \texttt{"visited\_url"}: "\url{https://m.163.com/dy/article/K30AB0GV0514R9P4.html}", \\
            \hspace*{1em} \texttt{"mode"}: "summarize", \\
            \hspace*{1em} \texttt{"result"}: \{ \\
            \hspace*{2em} \texttt{"rational"}: "The user seeks to identify a key figure in the American Revolutionary War and the year they authorized formal espionage activities. The content highlights John Jay's pivotal role in establishing counterintelligence and mentions Washington's 1776 correspondence about British spies. The formation of Jay's committee in 1776 directly addresses the user's goal.",    \\
            \hspace*{2em} \texttt{"evidence"}: [ \\
            \hspace*{2em} "华盛顿能够用谍获情，就知道反间工作同样必不可少。1776年3月24日，华盛顿在与部属交流的信中写道：‘我最担心的一件事，是他们的间谍，因此，我希望能对他们保持最严密的监视。’于是，浩浩荡荡的清除英国间谍活动在大陆军中展开。",
        "1776年6月，年仅30岁的约翰·杰伊受华盛顿的委派成立了‘纽约州侦查和挫败阴谋委员会’，负责情报搜集、逮捕英国间谍和调查可疑的英国同情者，拥有逮捕、审判、监禁、驱逐嫌疑犯的权力，约翰·杰伊出任该委员会主席至1777年2月中旬。"    \\
        \hspace*{2em} ], \\
            \hspace*{2em} \texttt{"summary"}: "在独立战争中，约翰·杰伊（John Jay）发挥了关键作用。1776年3月24日，华盛顿在信中表达了对英国间谍的担忧，随后于同年6月授权杰伊成立‘纽约州侦查和挫败阴谋委员会’，这是美国首次正式开展反间谍活动的机构。该委员会赋予杰伊逮捕、审判和驱逐英国间谍及同情者的权力，标志着美国早期情报工作的制度化。"  \\
            \}  \\
            \textbf{<\textbackslash tool\_response>} \\
    \end{stepTitle}
    \begin{stepTitle}{\textbf{Step 4:}}
        \textcolor{liautoblue}{
        \textbf{<think>} \\
        根据工具返回的结果，我已经找到了问题的答案。拉什莫尔山上雕刻的四位美国总统是乔治·华盛顿、托马斯·杰斐逊、西奥多·罗斯福和亚伯拉罕·林肯。在独立战争中发挥关键作用的人物是约翰·杰伊（John Jay）。他在1776年被授权组建“纽约州侦查和挫败阴谋委员会”，这是美国首次正式开展反间谍活动的机构。因此，我可以给出最终答案。\\
        \textbf{<\textbackslash think>} \\}
    \end{stepTitle}
    \textcolor{textred}{
        \textbf{<answer>} \\
            拉什莫尔山上雕刻的四位美国总统是乔治·华盛顿、托马斯·杰斐逊、西奥多·罗斯福和亚伯拉罕·林肯。在独立战争中发挥关键作用的人物是约翰·杰伊（John Jay）。他在1776年被授权组建“纽约州侦查和挫败阴谋委员会”，这是美国首次正式开展反间谍活动的机构。\\
        \textbf{<\textbackslash answer>}
    }
\end{toolbox}
\begin{toolbox}{Case3}
    \begin{center}
    \includegraphics[width=0.5\linewidth]{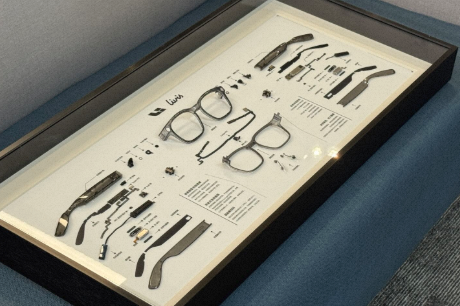}   \\
    \end{center}
    
    \textbf{Question:} What are the functions of these glasses? \\
    \textbf{Ground-truth:} Camera； Open-back headphones； Vehicle control； AI assistant\\
    \begin{stepTitle}{\textbf{Step 1:}}
        \textcolor{liautoblue}{
            \textbf{<think>} \\
            The image shows a disassembled pair of glasses with various components labeled. To determine the functions of these glasses, I need to identify each component and understand their roles. However, the labels on the components are not clearly visible in the current image. I will zoom in on the image to get a better view of the labels and components.
            \textbf{<\textbackslash think>} \\
        }
        \textbf{<tool\_call>}    \\
        \{
        "name": "im\_zoom\_in",
        "arguments": \{
            "bbox\_2d": [485, 385, 725, 580]
        \}
        \}  \\
        \textbf{<\textbackslash tool\_call>} \\  
        \textbf{<tool\_response>}    \\
        <image> \\
        \includegraphics[width=0.25\linewidth] {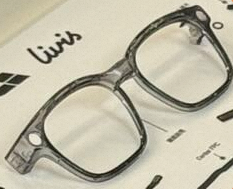}   \\
        \textbf{<\textbackslash tool\_response>} \\
    \end{stepTitle}
    \begin{stepTitle}{\textbf{Step 2:}}
        \textcolor{liautoblue}{
        \textbf{<think>} \\
        The zoomed-in image provides a clearer view of the glasses and some of their components. However, the labels on the components are still not legible. To understand the functions of these glasses, I need to identify the components and their roles. Since the labels are not readable, I will perform a web search using the brand name "Livis" to find more information about these glasses and their functionalities.\\
        \textbf{<\textbackslash think>} \\
        }
        \textbf{<tool\_call>}    \\
        \{
        "name": "web\_search",
        "arguments": \{
            "query": 'Livis glasses functions'
        \}
        \}  \\
        \textbf{<\textbackslash tool\_call>} \\
        \textbf{<tool\_response>}    \\
        \{ \\
        \hspace*{1em} \texttt{'search\_query'}: 'Livis glasses functions', \\
        \hspace*{1em} \texttt{'search\_result'}: [ \\
        \hspace*{2em} \{ \\
        \hspace*{3em} \texttt{'id'}: 1, \\
        \hspace*{3em} \texttt{'title'}: '视频评测｜理想 Livis：首先是副好眼镜，这真的很重要', \\

    \end{stepTitle}

    \end{toolbox}  
\begin{toolbox}{Case3}
     \begin{stepTitle}{}
        \hspace*{3em} \texttt{'content'}: '而为了让 Livis 作为一副眼镜更好用，理想也在镜片上花了不少功夫。理想 Livis 的镜片合作方为百年光学大厂蔡司...', \\
        \hspace*{3em} \texttt{'url'}: 'https://new.qq.com/rain/a/LNK2025120406148900', \\
        \hspace*{3em} \texttt{'date'}: '2025-12-04T12:05:00+08:00' \\
        \hspace*{2em} \}, \\
        \hspace*{2em} \dots \\
        \hspace*{2em} \{ \\
        \hspace*{3em} \texttt{'id'}: 5, \\
        \hspace*{3em} \texttt{'title'}: '1699 元！今年最值得买的智能眼镜 Livis，居然来自一家车企', \\
        \hspace*{3em} \texttt{'content'}: '依靠免唤醒和意图识别，你可以在任何地方直接说打开后备箱、查询胎压或者打开座椅加热，完全不需要掏出手机...', \\
        \hspace*{3em} \texttt{'url'}: 'https://new.qq.com/rain/a/LNK2025120312421600'' \\
        \hspace*{3em} \texttt{'date'}: '2025-12-03T22:21:00+08:00' \\
        \hspace*{2em} \} \\
        \hspace*{1em} ] \\
        \} \\
        \textbf{<\textbackslash tool\_response>} \\
     \end{stepTitle}

    \begin{stepTitle}{\textbf{Step 3:}}
        \textcolor{liautoblue}{
            \textbf{<think>} \\
            I am now ready to provide the final answer based on the information gathered. \\
            \textbf{<\textbackslash think>} \\
        }
    \end{stepTitle}

    \textcolor{textred}{
    \textbf{<answer>} \\
    The Livis glasses have multiple functions, including: \\
    1. **Optical Function**: High-quality lenses made in collaboration with Carl Zeiss, offering a wide range of vision correction options.\\
    2. **Sound System**: Sophisticated dual magnetic circuit sandwich speakers with intelligent frequency adjustment for leak-proof functionality and enhanced audio experience.\\
    3. **Camera Function**: A 12-megapixel IMX681 sensor with a 105-degree ultra-wide-angle lens, supporting flexible multi-format shooting and live photo capture.\\
    4. **Vehicle Control**: Voice-controlled vehicle operations, including opening the trunk, checking tire pressure, and adjusting seat heating.\\
    5. **Voice Assistant**: Integrated smart voice assistant for hands-free communication and access to various services.\\
    6. **Wireless Charging**: Dedicated charging case that supports wireless charging.\\
    7. **AI Features**: Advanced AI technology for recognizing user intentions, enhancing audio quality, and optimizing user interactions.\\
    These features make Livis glasses a multifunctional device that combines eyewear, audio, camera, vehicle control, and AI capabilities.\\
    \textbf{<\textbackslash answer>} \\
    }
\end{toolbox}

\newpage
\subsection{Prompt Design}
In this section, we display the prompts utilized by policy model and evaluation judge model.
\begin{toolbox}{Prompt: Policy Model}
     你是一个ReAct范式的agent，能够接受文本输入，回答用户问题对于一些复杂的问题，你可以选择调用网络搜索工具帮助你解决问题你可以调用的工具包括：\\
     \hspace*{1em} (**具体描述请参考工具描述部分**) \\
     \hspace*{1em} 1. Region Croping/Zooming \\
     \hspace*{1em} 2. Object Grounding Visual Search \\
     \hspace*{1em} 3. External Text Retrieval \\
     \hspace*{1em} 4. Webpage Content Retrieval \\
     \hspace*{1em} \dots \\
    对于每一个问题，你需要先思考，然后调用工具（如果需要），你会得到工具调用返回的结果，还可以根据工具的返回结果进行进一步的思考，最后给出答案你的思考过程，工具调用请求以及回答需要严格按照以下格式：\\
    <\textbackslash think> \\ 
    你的思考过程 \\
    <\textbackslash \/think> \\ 
    <\textbackslash tool\_call> \\
    {"name": <function-name>, "arguments": <args-json-object>} \\
    <\textbackslash \/tool\_call> (如果需要调用工具,你的工具调用请求参考usage中的示例) \\
    <\textbackslash think> \\
    你的思考过程 \\
    <\textbackslash \/think> (如果需要进一步思考) \\
    <\textbackslash answer> \\
    你的最终答案 \\
    <\textbackslash \/answer> \\
    请记住，你在每次调用工具之后，也就是输出</tool\_call>之后，都需要结束本轮对话，等待工具调用的结果返回，再进行后续动作在输出回答之后，即在输出<\/answer>之后，你需要立即结束本轮对话，不要再输出任何内容你的思考次数和工具调用次数没有限制，但必须在最后给出你的答案对于任何问题，你不应该拒绝回答，而应该通过不断思考或调用工具，直到得到确信的结果。
\end{toolbox}

\begin{toolbox}{Prompt: Evaluation Judge Model}
    你是一个公正的评测员，负责判断【模型回答】是否在事实和逻辑上符合【标准答案】。\\
    输入信息：\\
    - **问题**：{question}\\
    - **模型回答**：{output}\\
    - **标准答案 (GT)**：{ground\_truth}\\
    请遵循以下**通用等价性原则**进行判断：\\
    1. **核心事实一致性**：\\
       - 忽略措辞、语序或详细程度的差异。只要大概语义一致，即为正确。\\
    2. **数学与单位等价性**：\\
       - **自动换算**：如果模型回答的数值单位与 GT 或 问题要求的单位不同，**必须先进行数学换算**再比较。\\
       - **精度容忍**：允许合理的精度误差，换算后的前两位有效数字正确就行。\\
       - **格式识别**：识别 "45,610" (带逗号) 和 "45610" 为同一个数。\\
    3. **跨语言与语义对齐**：\\
       - 忽略中英文差异。 \\
       - 识别同义指代（例如 "NYC" = "New York City"; "巴特勒码头" = "Butler's Wharf"）。 \\
    4. **非法问题处理**：\\
       - 如果 GT 表示问题无解（如 "invalid question", "no answer"），只要模型**否定了事件发生**或**拒绝回答**，均判定为**正确**。\\
       - 只有当模型顺着错误前提编造事实时，才判错。\\
    5. **拒答处理**：\\
       - 如果 GT 有明确答案，而模型回答“不知道”或“无法回答”，判定为错误。\\
\end{toolbox}

\begin{toolbox}{Prompt: Evaluation Judge Model}

    Respond strictly in a valid JSON object matching this schema: \\
    \{ \\
    \hspace*{1em} \texttt{'extracted\_final\_answer'}: string, \\
    \hspace*{1em} \texttt{'reasoning'}: string, \\
    \hspace*{1em} \texttt{'result'}: "1" | "0", \\
    \hspace*{1em} \texttt{'confidence'}: integer, \\
    \hspace*{1em} \texttt{'strict'}: true \\
    \}
\end{toolbox}


\newpage

\bibliographystyle{plainnat}

\end{document}